%% file: main.tex
\pdfoutput=1
\documentclass[11pt,natbib]{article}

\usepackage[preprint]{coling}

\usepackage{times}
\usepackage{latexsym}

\usepackage[T1]{fontenc}

\usepackage[utf8]{inputenc}

\usepackage{microtype}

\usepackage{inconsolata}

\usepackage{graphicx}

\usepackage{algorithm}
\usepackage{algorithmic}
\usepackage{booktabs}
\usepackage{multirow}
\usepackage{amsmath}
\usepackage{newfloat}
\usepackage{listings}
\usepackage{graphicx}

\title{From Prediction to Application:\\Language Model-based Code Knowledge Tracing with Domain Adaptive Pre-Training and Automatic Feedback System with Pedagogical Prompting for Comprehensive Programming Education}

\author{
 \textbf{Unggi Lee\textsuperscript{1,6$*$}},
 \textbf{Jiyeong Bae\textsuperscript{1}},
 \textbf{Yeonji Jung\textsuperscript{2$\dagger$}},
 \textbf{Minji Kang\textsuperscript{3}},
 \textbf{Gyuri Byun\textsuperscript{4}},
 \textbf{Yeonseo Lee\textsuperscript{5}}
\\
 \textbf{Dohee Kim\textsuperscript{1}},
 \textbf{Sookbun Lee\textsuperscript{1}},
 \textbf{Jaekwon Park\textsuperscript{1}},
 \textbf{Taekyung Ahn\textsuperscript{1}},
 \textbf{Gunho Lee\textsuperscript{1}},
 \textbf{Hyeoncheol Kim\textsuperscript{6$\dagger$}}
\\
 Enuma, Inc.\textsuperscript{1},
 University of Memphis\textsuperscript{2},
 Daegu National University of Education\textsuperscript{3}
\\
 Seoul National University\textsuperscript{4},
 Seoul Metropolitan Office of Education\textsuperscript{5}
 Korea University\textsuperscript{6}
\\
 \small{
   \textbf{First Author$^*$ and Correspondences$^\dagger$:} \href{mailto:codingchild@korea.ac.kr}{codingchild@korea.ac.kr}, \href{mailto:yeonji.jung@memphis.edu}{yeonji.jung@memphis.edu}, \href{mailto:harrykim@korea.ac.kr}{harrykim@korea.ac.kr}
 }
}

\begin{document}
\maketitle

\input{0_abstract}

\input{1_introduction}
\input{2_related_work}

\input{3_method}

\input{4_result}

\input{5_conclusion}


\input{6_appendices}



\end{document}

%% file: 0_abstract.tex
\begin{abstract}
Knowledge Tracing (KT) is a critical component in online learning, but traditional approaches face limitations in interpretability and cross-domain adaptability. This paper introduces Language Model-based Code Knowledge Tracing (CodeLKT), an innovative application of Language model-based Knowledge Tracing (LKT) to programming education. CodeLKT leverages pre-trained language models to process learning data, demonstrating superior performance over existing KT and Code KT models. We explore Domain Adaptive Pre-Training (DAPT) and Task Adaptive Pre-Training (TAPT), showing enhanced performance in the coding domain and investigating cross-domain transfer between mathematics and coding. Additionally, we present an theoretically-informed integrated system combining CodeLKT with large language models to generate personalized, in-depth feedback to support students' programming learning. This work advances the field of Code Knowledge Tracing by expanding the knowledge base with language model-based approach and offering practical implications for programming education through data-informed feedback.
\end{abstract}

%% file: 1_introduction.tex
\section{Introduction}

\begin{figure*}[hbt!]
    \centering
    \includegraphics[width=\linewidth]{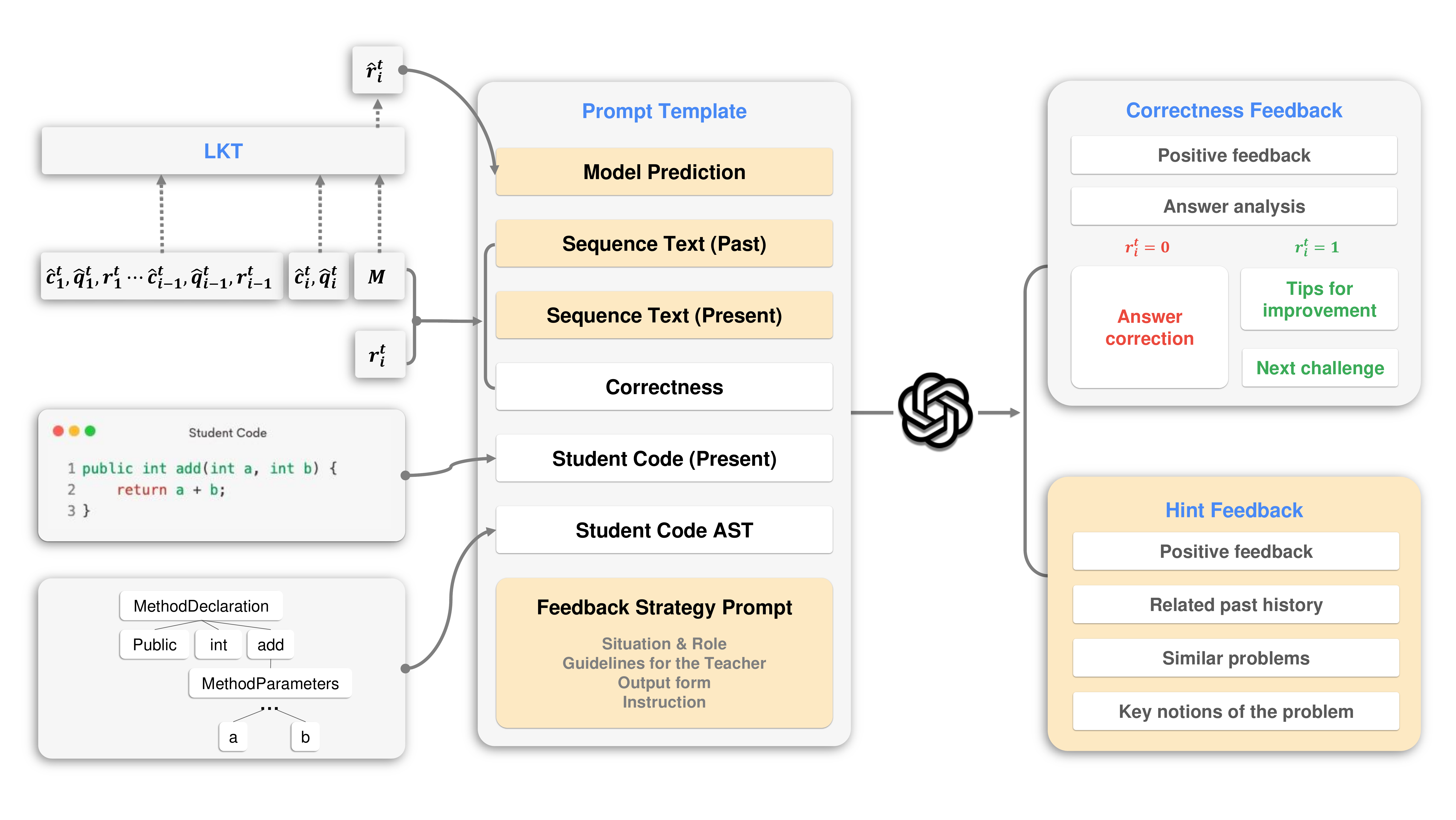}
    \caption{
    The prompt template for correctness and hint feedback consist of 7 and 4 components each; since hint feedback is given in the case that the student did not submit the answer, it does not contains 'Correctness', 'Student Code (Present)', and 'Student Code AST' components, while correctness feedback contains those. The correctness feedback provides answer correction when the student submitted wrong answer, and provides 'Tips for improvement' and 'Next challenge' in the opposite case. The both cases of correctness feedback contains 'Positive feedback', 'Answer analysis' and 'Comments for cheering up'. The hint feedback contains 'Positive feedback', 'Related past history', 'Similar problems' and 'Key notions of the problem'.}
    \label{fig:whole_arch}
\end{figure*}

In the rapidly evolving landscape of online programming education, Knowledge Tracing (KT) has emerged as a critical tool for modeling and predicting students' knowledge states over time \cite{piech2015deep}. However, as the demand for programming education grows, traditional KT approaches reveal significant limitations. Most existing KT models rely on sequences of numerical data, failing to capture the rich semantic information inherent in KT tasks \cite{liu2019ekt}. This limitation not only neglects the development of self-supervised learning and pre-training methods but also makes it challenging to transfer knowledge across domains or create foundation models for KT \cite{lee2024language}.

The field of Code Knowledge Tracing (CodeKT), which focuses specifically on modeling student knowledge in programming, lags even further behind \cite{shi2022code}. Despite the prevalence and importance of online programming education in today's digital-centric world \cite{kasurinen2009estimating, meliana2018adopting}, CodeKT has not fully adopted the methodological advances seen in general KT research.

Moreover, many current KT and CodeKT models focus solely on prediction, neglecting the crucial aspect of providing actionable insights and interventions. This narrow focus limits the practical utility of these models in real educational settings, where timely and targeted feedback can significantly enhance learning outcomes. The lack of integration between predictive models and feedback system applications represents a significant research gap in the field of Learning Analytics (LA).

To address these challenges, our research introduces Language Model-based Code Knowledge Tracing (CodeLKT), a novel approach inspired by Language Model-based Knowledge Tracing (LKT), that leverages the power of pre-trained language models. By utilizing natural language of student sequence data, CodeLKT aims to capture the nuanced semantics of programming tasks and student responses. This approach not only promises to enhance prediction accuracy but also opens new avenues for generating meaningful, context-aware feedback.

Our work also explores the effectiveness of Domain Adaptive Pre-Training (DAPT) and Task Adaptive Pre-Training (TAPT) in the coding domain. We investigate how these techniques can improve model performance in Code Knowledge Tracing tasks. Additionally, we examine the potential for knowledge transfer between related domains, particularly focusing on the interplay between mathematics and coding. This exploration provides valuable insights into the generalizability of language models (LMs) across different educational contexts and offers new perspectives on interdisciplinary learning.

We extend this prediction work to the design of implementing personalized, in-depth feedback to support students' programming skill. We present an innovative integrated system that combines CodeLKT's predictive capabilities with large language models (LLMs) to generate automated, pedagogically-sound feedback. This system leverages advanced prompting techniques grounded in programming education and pedagogical feedback theory to provide personalized and contextually relevant guidance to learners \cite{rivers2016learning}. 

By connecting prediction to learning application within a single framework, our approach represents a significant advancement towards more comprehensive and actionable programming education. This integration of performance prediction with tailored feedback generation addresses a critical gap in current educational technology, offering a more holistic solution for supporting student learning in programming courses. Our research contributions are below:

\begin{itemize}

\item \textbf{Introduction of CodeLKT:} We propose CodeLKT, a novel approach that significantly outperforms existing KT and Code KT models.

\item \textbf{Demonstration in effectiveness of Domain and Task Adaptation in the code domain:} We demonstrate the efficacy of DAPT and TAPT in the code domain. Our results show consistent performance improvements. We also explore the potential for knowledge transfer between mathematics and coding domains. Our findings reveal that models adapted to the mathematics domain perform well on CodeLKT tasks.

\item \textbf{Integrated Prediction-Application Framework:} We propose a novel framework that links prediction to intervention in programming education. By combining LLM with pedagogical prompting, our system not only predicts student performance but also provides tailored, theory-based feedback, advancing the field towards more comprehensive and actionable LA in programming education. 

\end{itemize}

%% file: 2_related_work.tex
\begin{table*}[hbt!]
\scriptsize
\centering
\begin{tabular}{clcccccc}
\toprule
\multirow{2}{*}{Type}   & \multicolumn{1}{c}{\multirow{2}{*}{Models}} & \multicolumn{2}{c}{CSEDM-19-Spring}                                   & \multicolumn{2}{c}{CSEDM-19-Fall}                                     & \multicolumn{2}{c}{CodeWorkout-Spr2019}                               \\
                        & \multicolumn{1}{c}{}                        & AUC                               & ACC                               & AUC                               & ACC                               & AUC                               & ACC                               \\
\toprule
LKT                     & BERT                                        & 0.8816±0.0329                     & 0.8990±0.0056                     & 0.8918±0.0050                     & 0.9028±0.0033                     & 0.8923±0.0106                     & 0.9017±0.0068                     \\
LKT                     & RoBERTa                                     & \textbf{0.9116±0.0096}                     & \textbf{0.9105±0.0096}                     & \textbf{0.9069±0.0031}                     & \textbf{0.9074±0.0028}                     & \underline{0.8985±0.0116}                     & \underline{0.9011±0.0105}                     \\
LKT                     & DistilBERT                                  & 0.8909±0.0078                     & 0.8965±0.0069                     & 0.8875±0.0050                     & 0.9001±0.0036                     & 0.8756±0.0264                     & 0.8942±0.0074                     \\
LKT                     & ALBERT                                      & 0.8053±0.1346                     & 0.8801±0.0218                     & 0.8047±0.0908                     & 0.8767±0.0150                     & 0.7823±0.1593                     & 0.8739±0.0291                     \\
LKT                     & ELECTRA                                     & 0.8697±0.0813                     & 0.8801±0.0218                     & 0.8564±0.0658                     & 0.8894±0.0240                     & 0.8558±0.0813                     & 0.8892±0.0312                     \\
LKT                     & ERNIE-2.0                                   & \underline{0.9005±0.0081}                     & \underline{0.9058±0.0081}                     & \underline{0.9051±0.0065}                     & \underline{0.9070±0.0020}                     & \textbf{0.8992±0.0088}                     & \textbf{0.9050±0.0077}                     \\
LKT                     & DeBERTa-v3                                  & 0.7587±0.0700                     & 0.8645±0.0149                     & 0.8461±0.0686                     & 0.8871±0.0225                     & 0.7490±0.0635                     & 0.8552±0.0078                     \\
DKT                     & DKT                                         & 0.7595±0.0117            & 0.8546±0.0145                     & 0.7721±0.0093                     & 0.8620±0.0056                     & 0.7477±0.0218                     & 0.8532±0.0273                     \\
DKT                     & DKVMN                                      & 0.7477±0.0218                     & 0.8532±0.0273                     & 0.7447±0.0208                     & 0.8522±0.0132                     & 0.7575±0.0210                     & 0.8524±0.0157                     \\
DKT                     & SAKT                                        & 0.7620±0.0033                     & 0.8920±0.0015                     & 0.7656±0.0079                     & 0.8720±0.0022                     & 0.7533±0.0140                     & 0.8534±0.0134                     \\
DKT                     & GKT (PAM)                                   & 0.7533±0.0140                     & 0.8534±0.0156                     & 0.7669±0.0134                     & 0.8608±0.0070                     & 0.7447±0.0208                     & 0.8522±0.0132                     \\
DKT                     & AKT                                         & 0.7601±0.0069                     & 0.8570±0.0147                     & 0.7713±0.0085                     & 0.8677±0.0068                     & 0.7485±0.0210                     & 0.8543±0.0270           \\
CodeDKT                     & CodeDKT                                         & 0.7431                     & -                     & -                     & -                     & -                     & -           \\
CodeDKT                     & ECKT                                         & 0.7653                     & -                     & -                     & -                     & -                     & -           \\
\toprule
\end{tabular}
\caption{Performance comparison of LKT and DKT models across three code-related datasets. Results are reported in AUC and ACC metrics. Note that the values for CodeDKT and ECKT are taken from their respective prior studies and are included for reference.}
\label{tb:performance}
\end{table*}

\section{Related Work}

\subsection{Code Knowledge Tracing}

The field of knowledge tracing in programming education has evolved significantly, starting with Bayesian Knowledge Tracing (BKT), which uses a Hidden Markov Model to track students' mastery of knowledge components based on their exercise performance \cite{kasurinen2009estimating, meliana2018adopting}. Despite its foundational role, BKT is limited in handling multi-skill exercises, prompting the development of models like the Additive Factor Model (AFM), which employs logistic regression to analyze multi-skill exercises through a Q-matrix, capturing student capability, KC difficulty, and learning rates \cite{rivers2016learning, hosseini2017stereotype}. DKT advanced this further by using RNN to predict future student performance based on past exercise sequences, though it faces challenges in interpretability due to the complexity of embeddings \cite{wang2017deep}.

Recent innovations have enhanced these models by incorporating detailed code analysis and leveraging large language models (LLMs). Code-DKT, for instance, improves traditional DKT by using an attention mechanism to extract domain-specific code features, thereby enhancing prediction accuracy \cite{shi2022code}. The Enhanced Code Knowledge Tracing (ECKT) framework further advances this approach by employing LLMs to generate detailed problem descriptions and knowledge concepts from student code through chain-of-thought prompting and few-shot learning. ECKT also integrates task difficulty information to provide a more nuanced assessment of student proficiency across various problem complexities \cite{yu2024eckt}.

\subsection{Domain Adaptative Pre-Training in Knowledge Tracing}

KT has seen several advancements in domain adaptation to address the challenge of limited student interaction data in new educational systems. Notable approaches include AdaptKT \cite{cheng2022adaptkt}, which uses instance selection and domain discrepancy minimization, Domain-Generalizable Knowledge Tracing (DGKT) \cite{xie2024domain} with its concept aggregation and relation-based attention, and Domain Adaptive Knowledge Tracing (DAKT) \cite{tang2024domain}, incorporating domain-shared answer embedding and adaptive knowledge state modeling. These methods have shown promise in enhancing model performance with limited training data in new domains.

However, traditional KT models have been constrained by their focus on numerical sequences for learning \cite{liu2019ekt}. This limitation has made it challenging to directly apply more advanced language model-based techniques that have proven effective in natural language processing tasks \cite{lee2024language, jung2024clst}.

In the field of natural language processing, techniques such as Domain Adaptive Pre-Training (DAPT) and Task Adaptive Pre-Training (TAPT) have emerged as powerful methods to improve model performance across different domains and tasks \cite{gururangan2020dont}. DAPT involves further pre-training of language models on domain-specific data, while TAPT focuses on task-specific data. These approaches allow models to better adapt to target domains or tasks \cite{singhal2023towards, wu2023bloom, labrak2024bio}.

\subsection{Automatic Feedback System for Programming Education}

Automatic feedback systems in programming education have garnered attention for their potential to enhance learning by providing immediate, actionable feedback. Such systems allow students to correct mistakes instantly and reinforce their understanding through iterative learning without waiting for instructor input \cite{rivers2016learning, keuning2018systematic}. This immediacy has been shown to improve retention of programming concepts and facilitate self-paced learning \cite{keuning2018systematic}.

Common feedback types include correctness feedback (which offers a binary assessment of code accuracy as an immediate validation that allows students to quickly adjust and retry) and hint feedback (which provides contextual clues to guide students toward correct solutions without revealing them outright) \cite{messer2024automated, keuning2018systematic}. These forms of feedback are essential for iterative learning and scaffolding, helping students refine their coding skills and develop problem-solving abilities \cite{cheng2023effects, keuning2018systematic}.

Despite these benefits, significant challenges remain, particularly in integrating individual students' learning status into meaningful scaffolding offered by automated feedback systems \cite{keuning2018systematic}. This issue is echoed by the current work on Knowledge Tracing (KT) and CodeKT models, which primarily focus on prediction \cite{shen2024survey, liu2022knowledge}. This narrow focus often ignores the crucial aspect of providing actionable insights that can be directly applied to learning practices, where timely feedback is essential for enhancing learning outcomes. 

To address these challenges, this study introduces an integrated approach that combines CodeLKT with large language models to generate personalized, pedagogically sound feedback in programming education. CodeLKT monitors the learner’s knowledge state over time, offering insights into their understanding and skill progression. By leveraging large language models, the system can generate correctness and hint feedback that is closely aligned with the learner’s current knowledge level and learning needs. This integration aims to expand codeLKT's implications, advancing both methodological and practical applications to enrich students' programming knowledge and learning experiences.

%% file: 3_method.tex
\section{Method}
\subsection{Code Language Model-based Knowledge Tracing}

\subsubsection{Problem Definition}

In the context of Knowledge Tracing (KT), the objective is to model and predict the knowledge state of students based on their interactions with programming educational content. Traditionally, this involves determining the likelihood that a student will correctly answer future questions based on their past responses. Formally, let $S = \{s_1, s_2, \ldots, s_N\}$ denote a set of students, and $C = \{c_1, c_2, \ldots, c_M\}$ denote a set of knowledge concepts (KCs), $Q = \{q_1, q_2, \ldots, q_M\}$ denote a set of questions. For student $s_j$, the interaction with a question is recorded as a tuple $(c_{ij}, q_{ij}, r_{ij})$, where $r_{ij} \in \{0, 1\}$ is the correctness indicator (1 if the answer is correct, 0 otherwise). The goal is to predict the correctness of future responses, $r_{ij}$, based on the sequence of past interactions.

\subsubsection{Language Model-based Code Knowledge Tracing}

We propose CodeLKT, a model that leverages pre-trained language models to capture semantic information from the textual content of programming KCs, questions, and responses. For a given sequence of interactions $(c_{1j}, q_{1j}, r_{1j}), (c_{2j}, q_{2j}, r_{2j}), \ldots, (c_{ij}, q_{ij}, r_{ij})$ for student $s_j$, each interaction is transformed into concatenated text as $(c_{1j}^t, q_{1j}^t, r_{1j}^t), (c_{2j}^t, q_{2j}^t, r_{2j}^t), \ldots, (c_{ij}^t, q_{ij}^t, r_{ij}^t)$. Here, $r_{ij}^t$ is represented by a special token: $\text{[CORRECT]}$ if $r_{ij}$ is 1, $\text{[INCORRECT]}$ if $r_{ij}$ is 0, and $\text{[MASK]}$ for predictions. 

The input $x_i$ of the model at each time step $i$ is formatted as below, where the $r_i^t$ has a $\text{[MASK]}$ token for prediction :

\begin{equation}
x_i = \text{[CLS]} \ c_1^t \ q_1^t \ r_1^t \ldots \ c_i^t \ q_i^t \ \text{[MASK]} \ \text{[SEP]}
\end{equation}

where $c_i^t$, $q_i^t$, and $r_i^t$ denote the text of KC, question, and response respectively.

The language model processes this sequence and outputs a hidden representation $h_{i}$ for each interaction. We then apply a linear transformation followed by a sigmoid function to predict the probability $\hat{r}_{i}$ of correctness for the next interaction:

\begin{equation}
\hat{r}_{i} = \sigma(W h_{i} + b)
\end{equation}

where $W$ and $b$ are trainable parameters, and $\sigma$ denotes the sigmoid function. The model is trained using binary cross-entropy loss:

\begin{equation}
\mathcal{L} = -\frac{1}{N} \sum_{i=1}^N \left( r_i \log(\hat{r}_i) + (1 - r_i) \log(1 - \hat{r}_i) \right)
\end{equation}

In summary, the LKT approach leverages the textual nature of programming questions and answers, allowing the language model to capture the underlying semantics and improve the accuracy of predicting future correctness.

\subsubsection{Textual Feature Extraction for Code Knowledge Tracing}
\subsubsection{Datasets}
\begin{figure*}[hbt!]
\centering
\includegraphics[width=\linewidth]{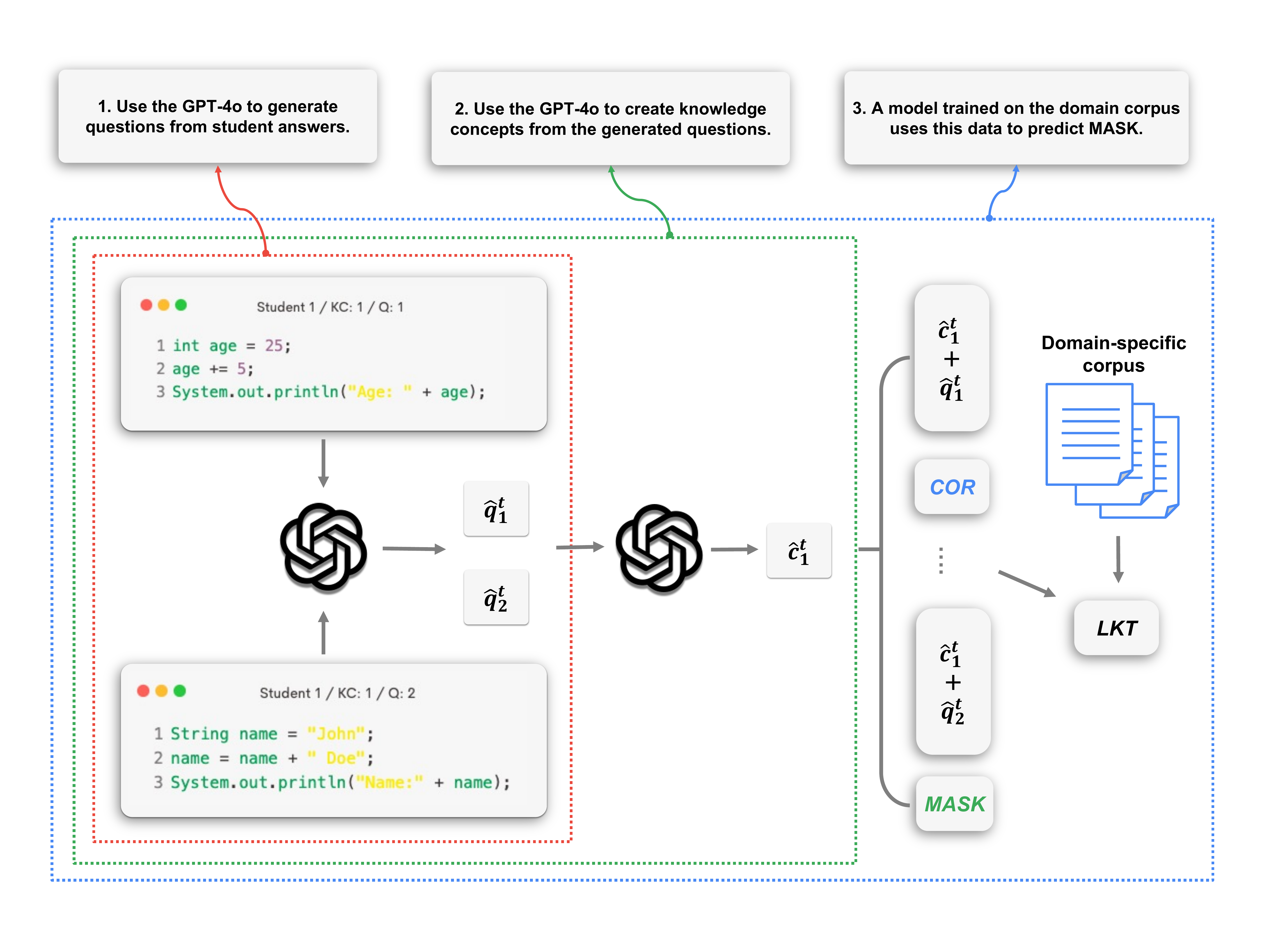}
\caption{
Pipeline to extract Question, Concept information for LKT from CSEDM-19-Spring, CSEDM-19-Fall, CodeWorkout-Spring2019 datasets. A model trained on the domain corpus uses this data to predict MASK.
}
\label{fig:data_gen}
\end{figure*}

When attempting to apply LKT to the Code dataset \cite{CSEDM2019}, we encountered a situation where student answers were provided as code-text, but the text for KCs and questions were only partially disclosed, with the rest being unavailable. However, each text was associated with the identifiers for the questions and KCs information. Therefore, we generated questions based on the students' code texts and created KCs from the collection of these generated questions, sequentially. (see Figure \ref{fig:data_gen})

\paragraph{Generate Questions from Answers}

Given a set of student answers, we use the GPT-4o \cite{openai2024hello} to generate questions that are no more than 200 characters long. Formally, for each answer of student $a_i^t$, we implement the LLM $\mathcal{L}$ to produce a corresponding question $\hat{q}_i^t$ with prompt text $p_q^t$:

\begin{equation}
\hat{q}_i^t = \mathcal{L}(a_i^t, p_q^t)
\end{equation}

\paragraph{Create Knowledge Concept Information}

The generated questions are then grouped into clusters based on their underlying KCs. Let $C = \{c_1^t, c_2^t, \ldots, c_i^t\}$ represent the set of knowledge KCs. Each question $\hat{q}_i^t$ is assigned to a KC $c_i^t$. We also implement the LLM $\mathcal{L}$ to produce a corresponding KC $\hat{c}_i^t$ with prompt text $p_c^t$:

\begin{equation}
\hat{c}_i^t = \mathcal{L}(\hat{q}_i^t, p_c^t)
\end{equation}

This process allows the LKT model to incorporate rich semantic information from the textual data, improving its ability to trace student knowledge and predict future performance accurately. The final form of CodeLKT dataset in this research, $x_i$ is below:

\begin{equation}
x_i = \text{[CLS]} \ \hat{c}_1^t \ \hat{q}_1^t \ r_1^t \ldots \ \hat{c}_i^t \ \hat{q}_i^t \ r_i^t \ \text{[SEP]}
\end{equation}


\subsubsection{Domain Adaptation}

We employed continual pre-training to adapt pre-trained language models to specific domains, such as Java, Python, and Math. The models are trained on domain-specific corpus using a Masked Language Modeling (MLM) objective, where 15\% of the tokens in the input sequence are randomly masked and the model learns to predict these masked tokens.



%

\subsection{Experiment Setup}
We conducted experiments to verify performance of CodeLKT, and effectiveness of DAPT and TAPT in CodeLKT.


\begin{itemize}
\item \textbf{LKT setting}: For our LKT (Learning Knowledge Tracing) setting, we used three code datasets and two math datasets: CSEDM-19-Spring, CSEDM-19-Fall, CodeWorkout-Spring2019 are code datasets, while DBE-KT22 and XES3G5M are math datasets. All code datasets were provided at the 2nd CSEDM workshop during LAK 2019 \cite{CSEDM2019}. DBE-KT22, from Australian National University \cite{abdelrahman2022dbe}, contains undergraduate Relational Databases course data. XES3G5M, created by TAL Education Group \cite{liu2024xes3g5m}, includes third-grade math performance data translated from Chinese to English.
\item \textbf{DAPT setting}: To implement domain adaptation, we used CodeXGLUE code2text-java \cite{cm_codexglue_code2text_java}, CodeXGLUE code2text-python \cite{cm_codexglue_code2text_python} as Java and Python corpus, and MetaMath \cite{yumetamath} as Mathematics corpus. The Java and Python corpus contain code data with pairing explanation of code. MetaMath contain question and answer about the math.
\end{itemize}

\subsubsection{Models}

\begin{itemize}

\item \textbf{Performance Comparison}: For performance comparison, we utilize LKTs and DKTs. Base models of LKTs are BERT \cite{devlin2018bert}, RoBERTa \cite{liu2019roberta}, DistilBERT \cite{sanh2019distilbert}, ALBERT \cite{lan2019albert}, ELECTRA \cite{clark2020electra}, ERNIE-2 \cite{sun2020ernie}, and DeBERTa-v3 \cite{he2021debertav3}, all of which are base-sized models. DKTs comprise DKT \cite{piech2015deep}, DKVMN \cite{zhang2017dynamic}, SAKT \cite{pandey2019self}, GKT \cite{nakagawa2019graph}, and AKT \cite{ghosh2020context}, which are representative models in the DKT category.

\item \textbf{DAPT and TAPT}: BERT serves as the comparison baseline. For code domain adaptation, we employ CodeBERT \cite{Feng2020CodeBERT}, BERT-Java-CL, and BERT-Python-CL. BERT-Java-CL is a BERT continually learned on the the Java corpus \cite{cm_codexglue_code2text_java}, while BERT-Python-CL is continually learned on the Python corpus \cite{cm_codexglue_code2text_python} dataset. BERT-MetaMath is continually trained BERT on the MetaMath dataset \cite{yumetamath}.
For code task adaptation, we utilize BERT-Spring-LKT, BERT-Fall-LKT, and BERT-Work-LKT, which are trained on code datasets using LKT techniques.
For math task adaptation, we employ BERT-XES-LKT and BERT-DBE-LKT, which are trained on math datasets using LKT techniques.
\item \textbf{Automatic Feedback System}: We used GPT-4o to create the automatic feedback system.

\end{itemize}

\begin{table*}[hbt!]
\scriptsize
\centering
\begin{tabular}{clcccccc}
\toprule
\multirow{2}{*}{Type} & \multicolumn{1}{c}{\multirow{2}{*}{Models}} & \multicolumn{2}{c}{CSEDM-19-Spring}                                   & \multicolumn{2}{c}{CSEDM-19-Fall}                                     & \multicolumn{2}{c}{CodeWorkout Spr2019}                               \\
                      & \multicolumn{1}{c}{}                        & AUC                               & ACC                               & AUC                               & ACC                               & AUC                               & ACC                               \\
\toprule
D                & BERT                                        & 0.8816±0.0329                     & 0.8990±0.0056                     & 0.8918±0.0050                     & 0.9028±0.0033                     & 0.8923±0.0106                     & 0.9017±0.0068                     \\
D                & CodeBERT                                    & \textbf{0.9107±0.0089} & \textbf{0.9083±0.0048} & \textbf{0.9033±0.0069} & \textbf{0.9079±0.0027} & \textbf{0.9085±0.0081} & \textbf{0.9071±0.0082} \\
D                & BERT-Java-CL                                & \underline{0.9008±0.0094}                     & 0.9018±0.0053                     & 0.8954±0.0070                     & 0.9042±0.0036                     & 0.8971±0.0117                     & 0.9016±0.0057                     \\
D                & BERT-Python-CL                              & 0.8917±0.0121                     & 0.9007±0.0031                     & 0.8975±0.0033                     & 0.9038±0.0013                     & 0.8936±0.0127                     & 0.9021±0.0061                     \\
D                & BERT-MetaMath                                & 0.8969±0.0073                     & \underline{0.9049±0.0069}                     & \underline{0.8980±0.0045}                     & \underline{0.9043±0.0033}                     & 0.8962±0.0120                     & 0.9043±0.0036                     \\

T                  & BERT-Spring-LKT                                & -                     & -                     & 0.8935±0.0073                     & 0.9029±0.0034                     & 0.8967±0.0133                     & \underline{0.9051±0.0077}                     \\

T                  & BERT-Fall-LKT                                & 0.8975±0.0053                     & 0.9039±0.0053                     & -                     & -                     & \underline{0.8987±0.0043}                     & 0.9049±0.0041                     \\

T                  & BERT-Work-LKT                                & 0.8989±0.0093                    & 0.9021±0.0038                     & 0.8946±0.0073                     & 0.9019±0.0055                     & -                     & -                     \\

T                  & BERT-XES-LKT                                & 0.8791±0.0156                     & 0.8943±0.0060                     & 0.8807±0.0050                     & 0.8981±0.0029                     & 0.8794±0.0087                     & 0.8950±0.0083                     \\
T                  & BERT-DBE-LKT                                & 0.8556±0.0411                     & 0.8851±0.0207                     & 0.8855±0.0057                     & 0.8997±0.0012                     & 0.8775±0.0130                     & 0.8972±0.0107  \\
\toprule
\end{tabular}
\caption{Comparison of DAPT and TAPT. All of the DAPT, including code and math, outperform the original BERT, especially CodeBERT is the best performance all of the dataset. In TAPT, code task adapted models outperform, but math task adapted models underperform.}
\label{tb:domain_task_adaptation}
\end{table*}

\subsubsection{Training and Evaluation}

We conducted our experiments using a 5-fold cross-validation approach to ensure robust performance estimation across different data splits. For training, we utilized a per-device batch size of 512, leveraging gradient accumulation. We employed Accelerate \cite{accelerate} for distributed training.

Domain adaptation was performed using BERT-Java-CL, BERT-Python-CL, and BERT-Math-CL models. We used a continual pre-training method, training on a corpus with a masking probability of 15\%, following the original BERT approach. These models were trained for up to 3 epochs.

For the LKT task, we fine-tuned the normal or domain adapted models. We used the AdamW \cite{loshchilov2019decoupledweightdecayregularization} optimizer with a learning rate of 5e-5 and weight decay of 0.01. The models were trained for up to 100 epochs, with early stopping implemented at a patience of 10 epochs.

Model performance was evaluated using Area Under the Curve (AUC) and Accuracy metrics. For each fold, we used the corresponding train and test splits. The final performance metrics were obtained by averaging the results across all five folds.

%% file: 4_result.tex
\section{Experimental Results}

\subsection{Performance of Comparison}

Table \ref{tb:performance} presents a comprehensive comparison of LKT and DKT models across three code-related datasets: CSEDM-19-Spring, CSEDM-19-Fall, and CodeWorkout Spr2019. The results clearly demonstrate the superiority of LKT models in Code Knowledge Tracing tasks. Among the LKT models, RoBERTa consistently outperforms others, achieving the highest AUC and ACC scores on both CSEDM datasets (AUC: 0.9116, 0.9069; ACC: 0.9105, 0.9074). ERNIE-2.0 follows closely, showing strong performance across all datasets and even surpassing RoBERTa on the CodeWorkout dataset (AUC: 0.8992, ACC: 0.9050).

In contrast, traditional DKT models and code-specific approaches like CodeDKT and ECKT demonstrate significantly lower performance. The best-performing DKT model, SAKT, achieves an AUC of only 0.7620 on CSEDM-19-Spring, substantially lower than the LKT models. Notably, CodeDKT and ECKT, despite being tailored for code-related tasks, report AUC scores of 0.7431 and 0.7653 respectively, which are considerably inferior to the LKT results. This performance gap underscores the remarkable effectiveness of language model-based approaches in capturing the intricacies of code-related knowledge tracing, marking a significant advancement over existing methodologies in this domain.

\subsection{DAPT and TAPT: Code Knowledge Tracing}
In this process, we examined not only the performance on Code data but also whether this performance improvement extends to another domain, mathematics.
Our first hypothesis questioned whether models adapted to the Code Corpus would perform better. We compared the performance of models with and without Domain Adaptation in CodeLKT scenarios. In addition, we investigated whether the type of programming language affected performance. For this experiment, we prepared an untrained BERT and three models adapted to the code domain: CodeBERT, BERT-Java-CL, and BERT-Python-CL. The latter two were specifically trained for this research using Continual Learning on Codexglue-code2text-java and Codexglue-code2text-python datasets, respectively.

The results (Table \ref{tb:domain_task_adaptation}) showed that all three models adapted to the code domain outperformed BERT across all datasets, indicating that domain adaptation using code corpus was effective. However, there was no consistent performance difference between BERT-Java-CL and BERT-Python-CL, suggesting that the specific programming language did not significantly impact performance, even though the csedm dataset is based on Java.

\subsection{DAPT and TAPT: Transfer Between Mathematics and Code Models}
Our second hypothesis explored the performance transfer between mathematics and code models (Table \ref{tb:domain_task_adaptation_math}). Previous studies have indicated a relationship between performance in mathematics and code tasks. We aimed to verify if this transfer occurs in LKT as well.
For the mathematics to code transfer, we used BERT-MetaMath, a model adapted to the mathematics domain using the MetaMath corpus. This model outperformed BERT on all CodeLKT datasets, demonstrating that mathematics Domain Adaptation was effective for CodeLKT. However, when we fine-tuned models with mathematics Task Adaptation (BERT-XES-LKT and BERT-DBE-LKT) on CodeLKT data, they underperformed compared to BERT, indicating that mathematics Task Adaptation was not effective for CodeLKT.
For the code to mathematics transfer, we trained CodeBERT, BERT-Java-CL, and BERT-Python-CL on mathematics LKT datasets (DBE-KT22, XES3G5M). These models outperformed the original BERT on all datasets, showing that models with code Domain Adaptation performed well on mathematics LKT. Similarly, models with code Task Adaptation (trained on BERT-Spring-LKT, BERT-Fall-LKT, BERT-Work-LKT) also outperformed the original BERT when trained on mathematics LKT datasets.
In conclusion, transfer between code and mathematics was effective in all cases except for mathematics Task Adaptation to CodeLKT. The reasons for this exception require further investigation in future research.

\begin{table}[hbt]
\centering
\scriptsize
\begin{tabular}{clcc}
\toprule
Type & \multicolumn{1}{c}{Models} & DBE-KT22                          & XES3G5M                           \\
\toprule
D    & BERT                       & 0.7452±0.0058                     & 0.8458±0.0011                     \\
D    & CodeBERT                   & \textbf{0.7963±0.0134}            & \textbf{0.8652±0.0008} \\
D    & BERT-Python-CL             & 0.7755±0.0122                     & 0.8574±0.0012                     \\
D    & BERT-Java-CL               & 0.7808±0.0073                     & \underline{0.8598±0.0016}                     \\
T    & BERT-Spring-LKT            & 0.7739±0.0042                     & 0.8572±0.0015                     \\
T    & BERT-Fall-LKT              & 0.7816±0.0030                     & 0.8580±0.0010                     \\
T    & BERT-Work-LKT              & \underline{0.7809±0.0075}         & 0.8564±0.0021 \\
\toprule
\end{tabular}
\caption{Comparison between DAPT and TAPT in Mathematics. All of the adapted models outperform comparing with original BERT.}
\label{tb:domain_task_adaptation_math}
\end{table}

\subsection{Results of Large Language Model-based Automatic Feedback System}
Beyond prediction to application, we designed an integrated system to generate personalized feedback based on the analysis of learners' knowledge status and history which is offered through a combination of CodeLKT and large language models (see Figure \ref{fig:whole_arch}). In alignment with the literature review \cite{messer2024automated, keuning2018systematic}, this system generates two types of feedback, depending on the timing of its application: correctness and hint feedback. Each type of feedback is designed to include the main four components, commonly used for feedback design in programming education \cite{shen2024survey, keuning2018systematic} (see Appendix 1 and 2). Both feedback types involve similar components (relating to students' learning history and providing positive feedback), but have different purposes for feedback with distinct components using different datasets (see Appendix 3). Correctness feedback focuses on helping learners first check whether their answers are correct ("Correction of the answer"), and then either identify areas for improvement with relevant guidance ("Analysis about the answer") or attempt higher levels of problems for those whose answers are correct ("Next challenge") (see Appendix 1 and 3). Hint feedback enables learners to refine their answers through generated hints ("Related past history", "Similar problems", and "Key notions of the problem") before submitting them to the system as a final one (see Appendix 2 and 3). 

To explore the potential effectiveness of our proposed feedback systems, we conducted a series of comparisons across three different approaches. Each comparison approach includes different prompts designed for the situation and role, guidelines for the teacher, learning history, output form, and instruction. The detailed results of these comparisons are included in Appendix 3 to 9, respectively. 
\begin{itemize}
\item \textbf{Comparison 1 (Proposed Approach)}: This method uses a comprehensive set of full prompts to guide how the GPT model should give feedback to the student (see Appendix 4 and 7). The dataset contained in the correctness feedback prompt includes LKT values, sequence texts of past problem and the past correctness, the present problem, the present student answer, its abstract syntax tree, and the correctness of the answer. In the case of hint feedback, the last three components, the present student answer, its abstract syntax tree, and the correctness of the answer, are eliminated. The past and present problems in the case of comparison 1 are inferred from the answer of the student by GPT. This combination allows us to provide highly personalized feedback based on the student's learning history, specific challenges, and current performance.

\item \textbf{Comparison 2 (Prompt-Only Method without LKT)}: This method simplifies our approach by removing LKT values and GPT-generated problems (see Appendix 5 and 8). Instead, we replace the model prediction with the DKT predicted probability of correctness, and the past and present problems with numerical data. The full guidance prompt and other components remain. While this method still offers designed feedback, it is less tailored, lacking the depth of context offered by the full LKT and GPT-inferred problem integration. 

\item \textbf{Comparison 3 (Basic Prompt-Only Approach)}: This approach simplifies the prompt to a single line: "You are a teacher who evaluates a student's programming skills and provides feedback." (see Appendix 6 and 9) It includes text-formed history, student code, and correctness but omits detailed prompts. As a result, the feedback is more generic and less tailored to the student’s needs.
\end{itemize}

%% file: 5_conclusion.tex
\section{Conclusion}

This paper introduced CodeLKT, a novel approach that significantly outperforms traditional KT and Code KT methods. By leveraging DAPT and TAPT, we demonstrated enhanced model performance and potential for knowledge transfer between related domains such as mathematics and coding. Our integrated prediction-application framework, combining CodeLKT with pedagogical prompting for automated feedback generation, addresses a critical gap in current educational technology.

While our results are promising, future work should focus on large-scale applications in diverse educational settings and investigation of long-term impacts on student learning outcomes. CodeLKT paves the way for more sophisticated, AI-driven educational tools in programming education, contributing to the advancement of both Code Knowledge Tracing and Learning Analytics. By bridging the gap between prediction and application, our work has the potential to significantly enhance the quality of online programming education and support more personalized learning experiences.

%% file: 6_appendices.tex
\newpage


\begin{table*}[htbp!]
\small
\centering
\caption*{\raggedright \large{\textbf{Appendix 1}}}
\begin{tabular}{p{3cm}p{5cm}p{8cm}}
\toprule
\textbf{Components} & \textbf{Prompt} & \textbf{Implemented Feedback} \\ 
\midrule
\textbf{Analysis about \newline the answer} & Consider the learning history of the student and provide personalized instruction to guide the code repair and improvement.
\newline - Analyse the learners's understanding based on the problem solved by the student in the past, problem that the student just solved, and the student's answer about the problem. \newline \newline \newline \newline \newline NO PROMPTS & 
- Your solution for replacing "zap" patterns with "zp" in a given string is an intelligent utilization of string manipulation methods. The way you've implemented logic using 'for' and 'if' statements to traverse through the given string, checking for the required pattern, and substituting it, is an excellent display of skills. ... This problem requires you to utilize skills you've developed during the string manipulations exercises, but also introduces you to new concepts. (*) \newline
- The use of the for loop, string-character check, concatenation using substrings, and the if condition for 'z' and 'p' is accurate and demonstrates your capacity to grasp the problem and translating it into comprehensive logic. ... Given that you have now successfully completed problem ['492', '34'], ... (**) \newline
- I appreciate the thought process behind your solution, using a loop to go through every character in the string and then replacing every instance of "z*p" with "zp". Your usage of the substring method to achieve this replacement is clever and shows a good grasp of Java's string methods. (***)
\\ 
\textbf{Correction of \newline the answer /  \newline Tips for \newline improvement} & Utilize corrective feedback that matches the developmental stage of the learner to lead corrective feedback to learning. \newline - Never include the full version of explicit correction or improved version of the code, but provide the partial sample code that is a corrected/improved version of the current answer of the student. \newline \newline \newline \newline \newline \newline \newline NO PROMPTS & 
- While your code works well for this particular problem, there might be a potential bottleneck. You're using the 'substring' method inside a loop which could slow down the execution for large inputs dramatically because string concatenation in Java results in creation of new String object. This might not be a problem right now, but... (*) \newline
- While your solution is correct, there's a bit of optimization you could do here. Modifying a string inside a loop can be costly because it creates a new String object each time. Instead, you ought to consider using a StringBuilder, which is more efficient in memory than a String when doing many concatenations.

Here's a short sample code with StringBuilder:

```java
StringBuilder result = new StringBuilder();

for (int i = 0; i < str.length() - 2; i++) {
    // your conditions here
    result.append(str.charAt(i));  // append instead of concatenation
}

return result.toString();" (**) \newline
- Do remember though, that replacing parts of a string using this approach can be inefficient in cases where the string is very large, because strings in Java are immutable. Any modification on a string will create a new copy. So each time through the loop you are creating a new string object. You might consider using a StringBuilder in those cases... (***)
\\ 
\textbf{Positive feedback} & Give positive feedback about the student's efforts to get the answer to address students' emotions and motivation by promoting the use of emotion control and motivating strategies. \newline \newline \newline \newline \newline NO PROMPTS & 
- Good job completing the advanced string manipulation challenge! I see a lot of progress since your last session, you have come a long way from making early mistakes to now providing impressive solutions. Your hard work and dedication is evident. (*) \newline
- Great job! Your efforts have paid off and you've got the correct answer! Your progress, especially given your history with concept 492, is commendable. Stay on this path and continue the good work. (**) \newline
- Great job on solving this problem! It seems like you have a good understanding of for loops, string manipulation and control flow with if statements. Your solution is efficient and meets the problem requirements perfectly. (***)
\\ 
\textbf{Next challenge} & Provide a problem of which the difficulty is a bit higher than the one that the student just solved, only in cases where the real result that the student got for this problem is 'Correct'. & 
- As you've advanced significantly in java string manipulations and concatenations, let's move to a new area that challenges your ability to implement string-based algorithms and data structures. ... This might seem complex at first, but remember to take it one step at a time. Start by splitting the string into words, then find a way to sort them based on their length, and finally combine them back into a single string. (*) \newline
- ... I'd recommend you take a step ahead and try solving a problem that requires the use of Java's Array objects. This will further enhance your ability to manipulate and iterate over different data types effectively. The problem ['506', '37'] would be a great start! (**)
\\ 
\bottomrule
\end{tabular}
\caption{Correctness Feedback Type Generation Prompts and Implemented Feedback. The symbols *, **, and *** indicate Comparison 1, Comparison 2, and Comparison 3 respectively.}
\label{tb:correctness_feedback1}
\end{table*}

\begin{table*}[htbp!]
\small
\centering
\caption*{\raggedright \large{\textbf{Appendix 2}}}
\begin{tabular}{p{3cm}p{5cm}p{8cm}}
\toprule
\textbf{Components} & \textbf{Prompt} & \textbf{Implemented Feedback} \\ 
\midrule
\textbf{Related past history} & 
- You should consider the student's language and programming level when you provide feedback. \newline
- You should consider the past history of the student when you give hints. For example, you can tell the student that the student has already solved similar problems before. \newline \newline \newline NO PROMPTS
 & 
- You have already solved a similar problem earlier where you manipulated two input strings by alternating characters and appending the extra characters in case one string was longer. You even wrote a function that created n repetitions of the last n characters of a string. So, I am confident that you are capable of resolving this pattern recognition problem as well. (*) \newline
- I noticed that you successfully solved question 33 which had the same concept ID ('492') as the current problem you're working on. Remember how you overcame the challenges on that problem? Let's use that same approach here. (**) \newline
- Think about how you can create a regular expression that matches 'zap' pattern. You may need to use string concatenation or string formatting to accomplish this. (***)
\\ 
\textbf{Positive feedback} & 
Give positive feedback about the student's efforts to get the answer to address students' emotions and motivation by promoting the use of emotion control and motivating strategies. \newline \newline \newline \newline NO PROMPTS & 
- I can see you're really putting in the effort to tackle these advanced string manipulation challenges. These are complex problems and it's great to see you persevere and show determination even when the tasks are difficult. Keep going! (*) \newline
- Great job on continuously working towards understanding the concepts! I know some of these problems can be challenging, but I also know that you are determined to overcome them. Every step, even mistakes, are a learning experience! (**) \newline
- Remember, practice makes perfect. Keep solving problems like these to improve your programming skills and your understanding of string manipulation. (***)
\\ 
\textbf{Similar problems} & 
Give hints to the student like showing similar problems. \newline \newline \newline \newline \newline \newline \newline \newline NO PROMPTS & 
- You can refer to the previous problem where you manipulated the string characters to form a new string. Similarly, in this problem, you need to detect a certain pattern and replace it. Use your previous knowledge to determine how to navigate the string, recognize the pattern and make the necessary replacements. (*) \newline
- A similar problem to this is the one mentioned above (question 33). This problem required you to apply the concept '492' in a similar way. Let's revisit that problem and try to draw parallels to the current problem. (**) \newline
- This problem involves replacing part of a string with a new string. Look into Python's `str.replace(old, new)` method or you can use Python's `re.sub(pattern, repl, string)` method from the regular expression (re) module. (***)
\\ 
\textbf{Key notions \newline of the problem} & 
Give hints to the student by presenting the key notions of the problem. \newline \newline \newline \newline \newline \newline \newline \newline \newline \newline \newline \newline \newline NO PROMPTS & 
- The key concept is 'string pattern recognition'. Here, the string pattern is 'zap', where 'a' can be any character. Your task is to replace this pattern with 'zp'. Think about how you can iterate over the string and identify the pattern. Also consider string replacement methods that might help. Lastly, remember to assemble the final string in the correct order after making all necessary replacements. Keep pushing, you're making solid progress! (*) \newline
- Notice that the concept '492' requires you to understand certain key principles in programming. While tackling this problem, focus particularly on these principles and strategies that you have used before in question 33. Let's carry your previous success into this new problem. Remember, the question is not whether you will understand it but when - because with your current pace, it's only a matter of time! Keep going! (**) \newline
- Note the pattern is 'zap' where 'a' can be any character. In other words, you are looking to replace "z(any character)p" with "zp". When using regular expressions, '.' is used to represent any character and '.*' is used to represent any number of characters. (***)
\\ 
\bottomrule
\end{tabular}
\caption{Hint Feedback Type Generation Prompts and Implemented Feedback. The symbols *, **, and *** indicate Comparison 1, Comparison 2, and Comparison 3 respectively.}
\label{tb:hint_feedback2}
\end{table*}

\begin{table*}[htbp!]
\small
\centering
\caption*{\raggedright \large{\textbf{Appendix 3}}}
\begin{tabular}{p{16cm}}
\toprule
\textbf{Data} \\ 
\midrule
\ A-1. The problem and correctness of each problem solved by the student in the past:\newline
\{Problem Text Past\} \newline
\newline
A-2. Problem and correctness of the problem that the student just solved: \newline
\{Problem Text Present\} \newline
\newline
A-3. The student's answer about the problem:   \newline
\{Response Code Present\}, \{Response Code AST\} \newline
\newline
A-4. The predicted probability  of the student getting this question correct:  \newline
\{Model Prob\} \newline
\newline
A-5. The real result that student got for this problem:  \newline
\{Correctness\} \newline
\newline
\newline
\newline
\ B-1. The concept IDs, question IDs, and correctness of each problem solved by the student in the past: \newline
\{Problem Past\} \newline
\newline
B-2. The concept IDs and question IDs that the student is solving: Problem and correctness of the problem that the student just solved: \newline
\{Problem Present\} \newline
\newline
B-3. The student answer about the problem:    \newline
\{Response Code Present\}, \{Response Code AST\} \newline
\newline
B-4. The predicted probability  of the student getting this question correct:   \newline
\{Model Prob\} \newline
\newline
B-5. The real result that student got for this problem: \newline
\{Correctness\} \newline
\newline
\\ 
\bottomrule
\end{tabular}
\caption{Correctness Feedback and Hint Feedback Type Generation Data.}
\label{tb:hint_feedback3}
\end{table*}

\begin{table*}[htbp!]
\small
\centering
\caption*{\raggedright \large{\textbf{Appendix 4}}}
\begin{tabular}{p{2cm}p{6.5cm}p{6.5cm}}
\toprule
\textbf{Case} & \textbf{Prompt} & \textbf{Student Answer} \\ 
\midrule
\textbf{Comparison 1} & 
\#\#\# Situation \& Role \newline
You are a teacher who evaluates a student's programming skills and provides feedback. The below outlines the part of learning history of the student you are tutoring and guidelines that you should consider as a teacher, conducting a one-on-one lesson. 
\newline \newline
\#\#\# Guidelines for the Teacher \newline
1. Consider the learning history of the student and provide personalized instruction to guide the code repair and improvement. \newline
- Analyse the learners's understanding based on the problem solved by the student in the past, a problem that the student just solved, and the student's answer about the problem. \newline
2. Utilize corrective feedback which matches with the developmental stage of the learner, in order to lead corrective feedback to learning. \newline
- Provide analysis about the answer code of the student with explicit code that the student submitted. \newline
- Never include the full version of explicit correction or improved version of the code, but provide the partial sample code that is a corrected/improved version of the current answer of the student. \newline
3. Give positive feedback about the student's efforts to get the answer to address students' emotions and motivation by promoting the use of emotion control and motivating strategies. \newline
4. Provide a problem of which the difficulty is a bit higher than the ""Problem that the student just solved"" only in cases of '5. The real result that student got for this problem' is 'Correct'. \newline
\newline
\#\#\# Learning history \newline
1. The problem and correctness of each problem solved by the student in the past:\newline
\{Problem Text Past\} \newline
\newline
2. Problem that the student just solved: \newline
\{Problem Text Present\} \newline
\newline
3. The student's answer about the problem: \newline
\{Response Code Present\}, \{Response Code AST\} \newline
\newline
4. The predicted probability  of the student getting this question correct: \newline
\{Model Prob\} \newline
\newline
5. The real result that student got for this problem: \newline
\{Correctness\} \newline
\newline
\#\#\# Output form: \newline
1. Positive feedback \newline
2. Analysis about the answer \newline
3. Correction of the answer / Tips for improvement \newline
4. Next challenge (provide this only in cases of '5. The real result that student got for this problem' is 'Correct')
comments for cheering up \newline
\newline
\#\#\# Instruction \newline
Refer to the past history, predicted probability of the student getting this question correct, and the real result that the student got for this problem, Provide appropriate feedback to the student in that situation. 
 & 
1. Positive feedback: \newline
   Good job completing the advanced string manipulation challenge! I see a lot of progress since your last session, you have come a long way from making early mistakes to now providing impressive solutions. Your hard work and dedication is evident. \newline
\newline
2. Analysis about the answer: \newline
   Your solution for replacing ""zap"" patterns with ""zp"" in a given string is an intelligent utilization of string manipulation methods. The way you've implemented logic using 'for' and 'if' statements to traverse through the given string, checking for the required pattern, and substituting it, is an excellent display of skills. You are proficiently laying out your control structures and syntax very well. \newline
\newline
3. Correction of the answer / Tips for improvement: \newline
   While your code works well for this particular problem, there might be a potential bottleneck. You're using the 'substring' method inside a loop which could slow down the execution for large inputs dramatically because string concatenation in Java results in creation of new String object. This might not be a problem right now, but it's beneficial to keep in mind the performance aspect while programming. As an improvement try using 'StringBuilder' or 'StringBuffer' implementations which can handle such operations more efficiently. \newline
\newline
4. Next challenge:
\newline
   As you've advanced significantly in java string manipulations and concatenations, let's move to a new area that challenges your ability to implement string-based algorithms and data structures. 
\newline \newline
   Here's a problem for you to solve:
   \newline
   Write a function that takes a string as input, which consists of words separated by spaces, and returns a new string where the words are sorted in the order of their length. If multiple words have the same length, they should retain their original order.
   \newline
   This might seem complex at first, but remember to take it one step at a time. Start by splitting the string into words, then find a way to sort them based on their length, and finally combine them back into a single string. This problem requires you to utilize skills you've developed during the string manipulations exercises, but also introduces you to new concepts. Good luck!
\\ 
\\ 
\bottomrule
\end{tabular}
\caption{Comparison 1: Prompts and Student Answer in Correctness Feedback Type}
\label{tb:hint_feedback4}
\end{table*}

\begin{table*}[htbp!]
\small
\centering
\caption*{\raggedright \large{\textbf{Appendix 5}}}
\begin{tabular}{p{2cm}p{6.5cm}p{6.5cm}}
\toprule
\textbf{Case} & \textbf{Prompt} & \textbf{Student Answer} \\ 
\midrule
\textbf{Comparison 2} & 
\#\#\# Situation \& Role \newline
You are a teacher who evaluates a student's programming skills and provides feedback. The below outlines the part of learning history of the student you are tutoring and guidelines that you should consider as a teacher, conducting a one-on-one lesson. 
\newline \newline
\#\#\# Guidelines for the Teacher \newline
1. Consider the learning history of the student and provide personalized instruction to guide the code repair and improvement. \newline
- Analyse the learners's understanding based on the problem solved by the student in the past, the problem that the student just solved, and the student's answer about the problem. \newline
2. Utilize corrective feedback which matches with the developmental stage of the learner, in order to lead corrective feedback to learning. \newline
- Provide analysis about the answer code of the student with explicit code that the student submitted. \newline
- Never include the full version of explicit correction or improved version of the code, but provide the partial sample code that is a corrected/improved version of the current answer of the student. \newline
3. Give positive feedback about the student's efforts to get the answer to address students' emotions and motivation by promoting the use of emotion control and motivating strategies. \newline
4. Provide a problem of which the difficulty is a bit higher than the ""Problem that the student just solved"" only in cases of '5. The real result that student got for this problem' is 'Correct'. \newline
\newline
\#\#\# Learning history \newline
1. The concept IDs, question IDs, and correctness of each problem solved by the student in the past:\newline
\{Problem Past\} \newline
\newline
2. The concept IDs and question IDs that the student is solving: \newline
\{Problem Present\} \newline
\newline
3. The student answer about the problem:  \newline
\{Response Code Present\}, \{Response Code AST\} \newline
\newline
4. The predicted probability  of the student getting this question correct: \newline
\{Model Prob\} \newline
\newline
5. The real result that student got for this problem:  \newline
\{Correctness\} \newline
\newline
\#\#\# Output form: \newline
1. Positive feedback \newline
2. Analysis about the answer \newline
3. Correction of the answer / Tips for improvement \newline
4. Next challenge (provide this only in cases of '5. The real result that student got for this problem' is 'Correct')
comments for cheering up \newline
\newline
\#\#\# Instruction \newline
Refer to the past history, predicted probability of the student getting this question correct, and the real result that the student got for this problem, Provide appropriate feedback to the student in that situation. 
 & 
1. Positive feedback: \newline
Great job! Your efforts have paid off and you've got the correct answer! Your progress, especially given your history with concept 492, is commendable. Stay on this path and continue the good work. 
 \newline
\newline
2. Analysis about the answer: \newline
Your interpretation of the problem and implementation of the solution display a good grasp of Java's string manipulation mechanisms and condition-based programming. The use of the for loop, string-character check, concatenation using substrings, and the if condition for 'z' and 'p' is accurate and demonstrates your capacity of grasping the problem and translating it into comprehensive logic.\newline
\newline
3. Correction of the answer / Tips for improvement: \newline
While your solution is correct, there's a bit of optimization you could do here. Modifying a string inside a loop can be costly because it creates a new String object each time. Instead, you ought to consider using a StringBuilder, which is more efficient in memory than a String when doing many concatenations. Note that this does not affect the correctness of your solution; it is just a suggestion to improve the performance.

Here's a short sample code with StringBuilder:

```java
StringBuilder result = new StringBuilder();

for (int i = 0; i < str.length() - 2; i++) {
    // your conditions here
    result.append(str.charAt(i));  // append instead of concatenation
}

return result.toString();
``` \newline
\newline
4. Next challenge:
\newline
Only provide this in cases of '5. The real result that student got for this problem' is 'Correct'. Given that you have now successfully completed problem ['492', '34'], I'd recommend you take a step ahead and try solving a problem that requires the use of Java's Array objects. This will further enhance your ability to manipulate and iterate over different data types effectively. The problem ['506', '37'] would be a great start! Keep pushing yourself!
\\ 
\\ 
\bottomrule
\end{tabular}
\caption{Comparison 2: Prompts and Student Answer in Correctness Feedback Type}
\label{tb:hint_feedback5}
\end{table*}

\begin{table*}[htbp!]
\small
\centering
\caption*{\raggedright \large{\textbf{Appendix 6}}}
\begin{tabular}{p{2cm}p{6.5cm}p{6.5cm}}
\toprule
\textbf{Case} & \textbf{Prompt} & \textbf{Student Answer} \\ 
\midrule
\textbf{Comparison 3} & 
\#\#\# Situation \& Role \newline
You are a teacher who evaluates a student's programming skills and provides feedback.
\newline \newline
\#\#\# Learning history \newline
1. The problem and correctness of each problem solved by the student in the past:\newline
\{Problem Text Past\} \newline
\newline
2. Problem that the student just solved: \newline
\{Problem Text Present\} \newline
\newline
3. The student answer about the problem:  \newline
\{Response Code Present\}, \{Response Code AST\} \newline
\newline
4. The predicted probability  of the student getting this question correct: \newline
\{Model Prob\} \newline
\newline
5. The real result that student got for this problem:  \newline
\{Correctness\} \newline
\newline
\#\#\# Instruction \newline
Provide appropriate feedback to the student in that situation.
 & 
Great job on solving this problem! It seems like you have a good understanding of for loops, string manipulation and control flow with if statements. Your solution is efficient and meets the problem requirements perfectly. 
 \newline
\newline
I appreciate the thought process behind your solution, using a loop to go through every character in the string and then replacing every instance of "z*p" with "zp". Your usage of the substring method to achieve this replacement is clever and shows a good grasp of Java's string methods.\newline
\newline
Do remember though, that replacing parts of a string using this approach can be inefficient in cases where the string is very large, because strings in Java are immutable. Any modification on a string will create a new copy. So each time through the loop you are creating a new string object. \newline
\newline
You might consider using a StringBuilder in those cases, which is mutable and allows modifications to the string it holds without creating new string objects. However, for a function of this size, your solution is fine. \newline
\newline
Keep up the good work!
\\ 
\\ 
\bottomrule
\end{tabular}
\caption{Comparison 3: Prompts and Student Answer in Correctness Feedback Type}
\label{tb:hint_feedback6}
\end{table*}

\begin{table*}[htbp!]
\small
\centering
\caption*{\raggedright \large{\textbf{Appendix 7}}}
\begin{tabular}{p{2cm}p{6.5cm}p{6.5cm}}
\toprule
\textbf{Case} & \textbf{Prompt} & \textbf{Student Answer} \\ 
\midrule
\textbf{Comparison 1} & 
\#\#\# Situation \& Role \newline
You are a teacher who evaluates a student's programming skills and provides feedback. The below outlines the part of learning history of the student you are tutoring and guidelines that you should consider as a teacher, conducting a one-on-one lesson.  
\newline \newline
\#\#\# Guidelines for feedback \newline
1. You should consider the student's language and programming level when you provide feedback. \newline
2. Give positive feedback about the student's efforts to get the answer to address students' emotions and motivation by promoting the use of emotion control and motivating strategies. \newline
3. Give hints to the student like showing similar problems and presenting the key notions of the problem. \newline
4. You should consider the past history of the student when you give hints. For example, you can tell the student that the student has already solved similar problems before. \newline
\newline
\#\#\# Learning history \newline
1. The problem and correctness of each problem solved by the student in the past:\newline
\{Problem Text Past\} \newline
\newline
2. Problem that the student just solved: \newline
\{Problem Text Present\} \newline
\newline
3. The predicted probability  of the student getting this question correct: \newline
\{Model Prob\} \newline
\newline
\newline
\#\#\# Output form: \newline
1. Positive feedback \newline
2. Related past history \newline
3. Similar problems \newline
4. Key notions of the problem
\newline
\newline
\#\#\# Instruction \newline
Refer to the past history, past problems that the student solved, and predicted probability of the student getting this question correct for this problem. Provide appropriate hints to the student in that situation.
 & 
1. Positive feedback \newline
I can see you're really putting in the effort to tackle these advanced string manipulation challenges. These are complex problems and it's great to see you persevere and show determination even when the tasks are difficult. Keep going!
 \newline
\newline
2. Related past history \newline
You have already solved a similar problem earlier where you manipulated two input strings by alternating characters and appending the extra characters in case one string was longer. You even wrote a function that created n repetitions of the last n characters of a string. So, I am confident that you are capable of resolving this pattern recognition problem as well.\newline
\newline
3. Similar problems \newline
You can refer to the previous problem where you manipulated the string characters to form a new string. Similarly, in this problem, you need to detect a certain pattern and replace it. Use your previous knowledge to determine how to navigate the string, recognize the pattern, and make the necessary replacements.\newline
\newline
4. Key notions of the problem
\newline
The key concept is 'string pattern recognition'. Here, the string pattern is 'zap', where 'a' can be any character. Your task is to replace this pattern with 'zp'. Think about how you can iterate over the string and identify the pattern. Also consider string replacement methods that might help. Lastly, remember to assemble the final string in the correct order after making all necessary replacements. Keep pushing, you're making solid progress!
\\ 
\\ 
\bottomrule
\end{tabular}
\caption{Comparison 1: Prompts and Student Answer in Hint Feedback Type}
\label{tb:hint_feedback7}
\end{table*}

\begin{table*}[htbp!]
\small
\centering
\caption*{\raggedright \large{\textbf{Appendix 8}}}
\begin{tabular}{p{2cm}p{6.5cm}p{6.5cm}}
\toprule
\textbf{Case} & \textbf{Prompt} & \textbf{Student Answer} \\ 
\midrule
\textbf{Comparison 2} & 
\#\#\# Situation \& Role \newline
You are a teacher who evaluates a student's programming skills and provides feedback. The below outlines the part of learning history of the student you are tutoring and guidelines that you should consider as a teacher, conducting a one-on-one lesson.  
\newline \newline
\#\#\# Guidelines for feedback \newline
1. You should consider the student's language and programming level when you provide feedback. \newline
2. Give positive feedback about the student's efforts to get the answer to address students' emotions and motivation by promoting the use of emotion control and motivating strategies. \newline
3. Give hints to the student like showing similar problems and presenting the key notions of the problem. \newline
4. You should consider the past history of the student when you give hints. For example, you can tell the student that the student has already solved similar problems before. \newline
\newline
\#\#\# Learning history \newline
1. The concept IDs, question IDs, and correctness of each problem solved by the student in the past:\newline
\{Problem Past\} \newline
\newline
2. The concept IDs and question IDs that the student is solving:\newline
\{Problem Present\} \newline
\newline
3. The predicted probability  of the student getting this question correct:  \newline
\{Model Prob\} \newline
\newline
\newline
\#\#\# Output form: \newline
1. Positive feedback \newline
2. Related past history \newline
3. Similar problems \newline
4. Key notions of the problem
\newline
\newline
\#\#\# Instruction \newline
Refer the past history, past problems that the student solved, predicted probability of the student getting this question correct for this problem. Provide appropriate hints to the student in that situation.
 & 
1. Positive feedback \newline
Great job on continuously working towards understanding the concepts! I know some of these problems can be challenging, but I also know that you are determined to overcome them. Every step, even mistakes, are a learning experience!
 \newline
\newline
2. Related past history \newline
I noticed that you successfully solved question 33 which had the same concept ID ('492') as the current problem you're working on. Remember how you overcame the challenges on that problem? Let's use that same approach here.\newline
\newline
3. Similar problems \newline
A similar problem to this is the one mentioned above (question 33). This problem required you to apply the concept '492' in a similar way. Let's revisit that problem and try to draw parallels to the current problem.\newline
\newline
4. Key notions of the problem
\newline
Notice that the concept '492' requires you to understand certain key principles in programming. While tackling this problem, focus particularly on these principles and strategies that you have used before in question 33. Let's carry your previous success into this new problem. Remember, the question is not whether you will understand it but when - because with your current pace, it's only a matter of time! Keep going!
\\ 
\\ 
\bottomrule
\end{tabular}
\caption{Comparison 2: Prompts and Student Answer in Hint Feedback Type}
\label{tb:hint_feedback8}
\end{table*}

\begin{table*}[htbp!]
\small
\centering
\caption*{\raggedright \large{\textbf{Appendix 9}}}
\begin{tabular}{p{2cm}p{6.5cm}p{6.5cm}}
\toprule
\textbf{Case} & \textbf{Prompt} & \textbf{Student Answer} \\ 
\midrule
\textbf{Comparison 3} & 
\#\#\# Situation \& Role \newline
You are a teacher who evaluates a student's programming skills and provides feedback.  
\newline \newline
\#\#\# Learning history \newline
1. The problem and correctness of each problem solved by the student in the past:\newline
\{Problem Text Past\} \newline
\newline
2. Problem that the student just solved:\newline
\{Problem Text Present\} \newline
\newline
3. The predicted probability  of the student getting this question correct: \newline
\{Model Prob\} \newline
\newline
\newline
\#\#\# Instruction \newline
Provide appropriate hints to the student in that situation.
 & 
To solve this problem, you need to focus on string manipulation methods. Here are some hints to consider: \newline
\newline
1. This problem involves replacing part of a string with a new string. Look into Python's `str.replace(old, new)` method or you can use Python's `re.sub(pattern, repl, string)` method from the regular expression (re) module.
 \newline
\newline
2. Note the pattern is 'zap' where 'a' can be any character. In other words, you are looking to replace ""z(any character)p"" with ""zp"". When using regular expressions, '.' is used to represent any character and '.*' is used to represent any number of characters.\newline
\newline
3. Think about how you can create a regular expression that matches 'zap' pattern. You may need to use string concatenation or string formatting to accomplish this.
\newline
\newline
4. Write down the steps of the problem and try to follow them while creating your solution. It is a good practice to first solve it manually for a given string and then code it for the generalized problem. 
\newline
\newline
Remember, practice makes perfect. Keep solving problems like these to improve your programming skills and your understanding of string manipulation.
\\ 
\\ 
\bottomrule
\end{tabular}
\caption{Comparison 3: Prompts and Student Answer in Hint Feedback Type}
\label{tb:hint_feedback9}
\end{table*}

%% file: main.bbl
@inproceedings{kour2014real,
  title={Real-time segmentation of on-line handwritten arabic script},
  author={Kour, George and Saabne, Raid},
  booktitle={Frontiers in Handwriting Recognition (ICFHR), 2014 14th International Conference on},
  pages={417--422},
  year={2014},
  organization={IEEE}
}

@inproceedings{kour2014fast,
  title={Fast classification of handwritten on-line Arabic characters},
  author={Kour, George and Saabne, Raid},
  booktitle={Soft Computing and Pattern Recognition (SoCPaR), 2014 6th International Conference of},
  pages={312--318},
  year={2014},
  organization={IEEE},
  doi={10.1109/SOCPAR.2014.7008025}
}

@inproceedings{ghosh2021option,
  title={Option tracing: Beyond correctness analysis in knowledge tracing},
  author={Ghosh, Arita and Raspat, Jay and Lan, Andrew},
  booktitle={International Conference on Artificial Intelligence in Education},
  pages={137--149},
  year={2021},
  month={June},
  publisher={Springer International Publishing},
  address={Cham}
}

@article{hadash2018estimate,
  title={Estimate and Replace: A Novel Approach to Integrating Deep Neural Networks with Existing Applications},
  author={Hadash, Guy and Kermany, Einat and Carmeli, Boaz and Lavi, Ofer and Kour, George and Jacovi, Alon},
  journal={arXiv preprint arXiv:1804.09028},
  year={2018}
}

@article{lan2019albert,
  title={Albert: A lite bert for self-supervised learning of language representations},
  author={Lan, Zhenzhong and Chen, Mingda and Goodman, Sebastian and Gimpel, Kevin and Sharma, Piyush and Soricut, Radu},
  journal={arXiv preprint arXiv:1909.11942},
  year={2019}
}

@article{devlin2018bert,
  title={Bert: Pre-training of deep bidirectional transformers for language understanding},
  author={Devlin, Jacob and Chang, Ming-Wei and Lee, Kenton and Toutanova, Kristina},
  journal={arXiv preprint arXiv:1810.04805},
  year={2018}
}

@article{sanh2019distilbert,
  title={DistilBERT, a distilled version of BERT: smaller, faster, cheaper and lighter},
  author={Sanh, Victor and Debut, Lysandre and Chaumond, Julien and Wolf, Thomas},
  journal={arXiv preprint arXiv:1910.01108},
  year={2019}
}

@article{liu2019roberta,
  title={Roberta: A robustly optimized bert pretraining approach},
  author={Liu, Yinhan and Ott, Myle and Goyal, Naman and Du, Jingfei and Joshi, Mandar and Chen, Danqi and Levy, Omer and Lewis, Mike and Zettlemoyer, Luke and Stoyanov, Veselin},
  journal={arXiv preprint arXiv:1907.11692},
  year={2019}
}

@article{zaheer2020big,
  title={Big bird: Transformers for longer sequences},
  author={Zaheer, Manzil and Guruganesh, Guru and Dubey, Kumar Avinava and Ainslie, Joshua and Alberti, Chris and Ontanon, Santiago and Pham, Philip and Ravula, Anirudh and Wang, Qifan and Yang, Li and others},
  journal={Advances in neural information processing systems},
  volume={33},
  pages={17283--17297},
  year={2020}
}

@article{he2021debertav3,
  title={Debertav3: Improving deberta using electra-style pre-training with gradient-disentangled embedding sharing},
  author={He, Pengcheng and Gao, Jianfeng and Chen, Weizhu},
  journal={arXiv preprint arXiv:2111.09543},
  year={2021}
}

@article{jiang2020convbert,
  title={Convbert: Improving bert with span-based dynamic convolution},
  author={Jiang, Zi-Hang and Yu, Weihao and Zhou, Daquan and Chen, Yunpeng and Feng, Jiashi and Yan, Shuicheng},
  journal={Advances in Neural Information Processing Systems},
  volume={33},
  pages={12837--12848},
  year={2020}
}

@article{clark2020electra,
  title={Electra: Pre-training text encoders as discriminators rather than generators},
  author={Clark, Kevin and Luong, Minh-Thang and Le, Quoc V and Manning, Christopher D},
  journal={arXiv preprint arXiv:2003.10555},
  year={2020}
}

@inproceedings{sun2020ernie,
  title={Ernie 2.0: A continual pre-training framework for language understanding},
  author={Sun, Yu and Wang, Shuohuan and Li, Yukun and Feng, Shikun and Tian, Hao and Wu, Hua and Wang, Haifeng},
  booktitle={Proceedings of the AAAI conference on artificial intelligence},
  volume={34},
  pages={8968--8975},
  year={2020}
}

@article{sun2021ernie,
  title={Ernie 3.0: Large-scale knowledge enhanced pre-training for language understanding and generation},
  author={Sun, Yu and Wang, Shuohuan and Feng, Shikun and Ding, Siyu and Pang, Chao and Shang, Junyuan and Liu, Jiaxiang and Chen, Xuyi and Zhao, Yanbin and Lu, Yuxiang and others},
  journal={arXiv preprint arXiv:2107.02137},
  year={2021}
}

@article{peng2021mathbert,
  title={Mathbert: A pre-trained model for mathematical formula understanding},
  author={Peng, Shuai and Yuan, Ke and Gao, Liangcai and Tang, Zhi},
  journal={arXiv preprint arXiv:2105.00377},
  year={2021}
}

@article{liu2024xes3g5m,
  title={XES3G5M: A Knowledge Tracing Benchmark Dataset with Auxiliary Information},
  author={Liu, Zitao and Liu, Qiongqiong and Guo, Teng and Chen, Jiahao and Huang, Shuyan and Zhao, Xiangyu and Tang, Jiliang and Luo, Weiqi and Weng, Jian},
  journal={Advances in Neural Information Processing Systems},
  volume={36},
  year={2024}
}

@article{abdelrahman2022dbe,
  title={DBE-KT22: A Knowledge Tracing Dataset Based on Online Student Evaluation},
  author={Abdelrahman, Ghodai and Abdelfattah, Sherif and Wang, Qing and Lin, Yu},
  journal={arXiv preprint arXiv:2208.12651},
  year={2022}
}

@article{achiam2023gpt,
  title={Gpt-4 technical report},
  author={Achiam, Josh and Adler, Steven and Agarwal, Sandhini and Ahmad, Lama and Akkaya, Ilge and Aleman, Florencia Leoni and Almeida, Diogo and Altenschmidt, Janko and Altman, Sam and Anadkat, Shyamal and others},
  journal={arXiv preprint arXiv:2303.08774},
  year={2023}
}

@inproceedings{yeung2018addressing,
  title={Addressing two problems in deep knowledge tracing via prediction-consistent regularization},
  author={Yeung, Chun-Kit and Yeung, Dit-Yan},
  booktitle={Proceedings of the fifth annual ACM conference on learning at scale},
  pages={1--10},
  year={2018}
}

@inproceedings{lee2019knowledge,
  title={Knowledge query network for knowledge tracing: How knowledge interacts with skills},
  author={Lee, Jinseok and Yeung, Dit-Yan},
  booktitle={Proceedings of the 9th international conference on learning analytics \& knowledge},
  pages={491--500},
  year={2019}
}

@inproceedings{zhang2017dynamic,
  title={Dynamic key-value memory networks for knowledge tracing},
  author={Zhang, Jiani and Shi, Xingjian and King, Irwin and Yeung, Dit-Yan},
  booktitle={Proceedings of the 26th international conference on World Wide Web},
  pages={765--774},
  year={2017}
}

@inproceedings{nakagawa2019graph,
  title={Graph-based knowledge tracing: modeling student proficiency using graph neural network},
  author={Nakagawa, Hiromi and Iwasawa, Yusuke and Matsuo, Yutaka},
  booktitle={IEEE/WIC/ACM International Conference on Web Intelligence},
  pages={156--163},
  year={2019}
}

@article{pandey2019self,
  title={A self-attentive model for knowledge tracing},
  author={Pandey, Shalini and Karypis, George},
  journal={arXiv preprint arXiv:1907.06837},
  year={2019}
}

@inproceedings{choi2020towards,
  title={Towards an appropriate query, key, and value computation for knowledge tracing},
  author={Choi, Youngduck and Lee, Youngnam and Cho, Junghyun and Baek, Jineon and Kim, Byungsoo and Cha, Yeongmin and Shin, Dongmin and Bae, Chan and Heo, Jaewe},
  booktitle={Proceedings of the seventh ACM conference on learning@ scale},
  pages={341--344},
  year={2020}
}

@inproceedings{ghosh2020context,
  title={Context-aware attentive knowledge tracing},
  author={Ghosh, Aritra and Heffernan, Neil and Lan, Andrew S},
  booktitle={Proceedings of the 26th ACM SIGKDD international conference on knowledge discovery \& data mining},
  pages={2330--2339},
  year={2020}
}

@article{lee2022monacobert,
  title={MonaCoBERT: Monotonic attention based ConvBERT for Knowledge Tracing},
  author={Lee, Unggi and Park, Yonghyun and Kim, Yujin and Choi, Seongyune and Kim, Hyeoncheol},
  journal={arXiv preprint arXiv:2208.12615},
  year={2022}
}

@article{abdelrahman2023knowledge,
  title={Knowledge tracing: A survey},
  author={Abdelrahman, Ghodai and Wang, Qing and Nunes, Bernardo},
  journal={ACM Computing Surveys},
  volume={55},
  number={11},
  pages={1--37},
  year={2023},
  publisher={ACM New York, NY}
}

@inproceedings{pandey2020rkt,
  title={RKT: relation-aware self-attention for knowledge tracing},
  author={Pandey, Shalini and Srivastava, Jaideep},
  booktitle={Proceedings of the 29th ACM International Conference on Information \& Knowledge Management},
  pages={1205--1214},
  year={2020}
}

@inproceedings{yang2021gikt,
  title={GIKT: a graph-based interaction model for knowledge tracing},
  author={Yang, Yang and Shen, Jian and Qu, Yanru and Liu, Yunfei and Wang, Kerong and Zhu, Yaoming and Zhang, Weinan and Yu, Yong},
  booktitle={Machine Learning and Knowledge Discovery in Databases: European Conference, ECML PKDD 2020, Ghent, Belgium, September 14--18, 2020, Proceedings, Part I},
  pages={299--315},
  year={2021},
  organization={Springer}
}

@inproceedings{liu2021improving,
  title={Improving knowledge tracing via pre-training question embeddings},
  author={Liu, Yunfei and Yang, Yang and Chen, Xianyu and Shen, Jian and Zhang, Haifeng and Yu, Yong},
  booktitle={Proceedings of the Twenty-Ninth International Conference on International Joint Conferences on Artificial Intelligence},
  pages={1556--1562},
  year={2021}
}

@inproceedings{su2018exercise,
  title={Exercise-enhanced sequential modeling for student performance prediction},
  author={Su, Yu and Liu, Qingwen and Liu, Qi and Huang, Zhenya and Yin, Yu and Chen, Enhong and Ding, Chris and Wei, Si and Hu, Guoping},
  booktitle={Proceedings of the AAAI Conference on Artificial Intelligence},
  volume={32},
  year={2018}
}

@article{liu2019ekt,
  title={Ekt: Exercise-aware knowledge tracing for student performance prediction},
  author={Liu, Qi and Huang, Zhenya and Yin, Yu and Chen, Enhong and Xiong, Hui and Su, Yu and Hu, Guoping},
  journal={IEEE Transactions on Knowledge and Data Engineering},
  volume={33},
  number={1},
  pages={100--115},
  year={2019},
  publisher={IEEE}
}

@inproceedings{lee2022contrastive,
  title={Contrastive learning for knowledge tracing},
  author={Lee, Wonsung and Chun, Jaeyoon and Lee, Youngmin and Park, Kyoungsoo and Park, Sungrae},
  booktitle={Proceedings of the ACM Web Conference 2022},
  pages={2330--2338},
  year={2022}
}

@inproceedings{nagatani2019augmenting,
  title={Augmenting knowledge tracing by considering forgetting behavior},
  author={Nagatani, Koki and Zhang, Qian and Sato, Masahiro and Chen, Yan-Ying and Chen, Francine and Ohkuma, Tomoko},
  booktitle={The world wide web conference},
  pages={3101--3107},
  year={2019}
}

@inproceedings{chen2017tracking,
  title={Tracking knowledge proficiency of students with educational priors},
  author={Chen, Yuying and Liu, Qi and Huang, Zhenya and Wu, Le and Chen, Enhong and Wu, Runze and Su, Yu and Hu, Guoping},
  booktitle={Proceedings of the 2017 ACM on Conference on Information and Knowledge Management},
  pages={989--998},
  year={2017}
}

@inproceedings{wang2021temporal,
  title={Temporal cross-effects in knowledge tracing},
  author={Wang, Chenyang and Ma, Weizhi and Zhang, Min and Lv, Chuancheng and Wan, Fengyuan and Lin, Huijie and Tang, Taoran and Liu, Yiqun and Ma, Shaoping},
  booktitle={Proceedings of the 14th ACM International Conference on Web Search and Data Mining},
  pages={517--525},
  year={2021}
}

@article{abdelrahman2022deep,
  title={Deep graph memory networks for forgetting-robust knowledge tracing},
  author={Abdelrahman, Ghodai and Wang, Qing},
  journal={IEEE Transactions on Knowledge and Data Engineering},
  year={2022},
  publisher={IEEE}
}

@inproceedings{shen2021learning,
  title={Learning process-consistent knowledge tracing},
  author={Shen, Shuanghong and Liu, Qi and Chen, Enhong and Huang, Zhenya and Huang, Wei and Yin, Yu and Su, Yu and Wang, Shijin},
  booktitle={Proceedings of the 27th ACM SIGKDD conference on knowledge discovery \& data mining},
  pages={1452--1460},
  year={2021}
}

@inproceedings{liu2023enhancing,
  title={Enhancing deep knowledge tracing with auxiliary tasks},
  author={Liu, Zitao and Liu, Qiongqiong and Chen, Jiahao and Huang, Shuyan and Gao, Boyu and Luo, Weiqi and Weng, Jian},
  booktitle={Proceedings of the ACM Web Conference 2023},
  pages={4178--4187},
  year={2023}
}


@article{JunyiOnlineLearningDataset,
  title={Junyi Academy Online Learning Activity Dataset: A large-scale public online learning activity dataset from elementary to senior high school students.},
  author={Pojen, Chen and Mingen, Hsieh and Tzuyang, Tsai},
  journal={Dataset available from https://www.kaggle.com/junyiacademy/learning-activity-public-dataset-by-junyi-academy},
  year={2020}
}


@article{bommasani2021opportunities,
  title={On the opportunities and risks of foundation models},
  author={Bommasani, Rishi and Hudson, Drew A and Adeli, Ehsan and Altman, Russ and Arora, Simran and von Arx, Sydney and Bernstein, Michael S and Bohg, Jeannette and Bosselut, Antoine and Brunskill, Emma and others},
  journal={arXiv preprint arXiv:2108.07258},
  year={2021}
}

@inproceedings{zhao2020cold,
  title={Cold start knowledge tracing with attentive neural turing machine},
  author={Zhao, Jinjin and Bhatt, Shreyansh and Thille, Candace and Gattani, Neelesh and Zimmaro, Dawn},
  booktitle={Proceedings of the Seventh ACM Conference on Learning@ Scale},
  pages={333--336},
  year={2020}
}

@inproceedings{das2021new,
  title={A New Interpretation of Knowledge Tracing Models' Predictive Performance in Terms of the Cold Start Problem.},
  author={Das, Rohini and Zhang, Jiayi and Baker, Ryan S and Scruggs, Richard},
  booktitle={EDM (Workshops)},
  year={2021}
}

@article{corbett1994knowledge,
  title={Knowledge tracing: Modeling the acquisition of procedural knowledge},
  author={Corbett, Albert T and Anderson, John R},
  journal={User modeling and user-adapted interaction},
  volume={4},
  pages={253--278},
  year={1994},
  publisher={Springer}
}

@article{dhawan2020online,
  title={Online learning: A panacea in the time of COVID-19 crisis},
  author={Dhawan, Shivangi},
  journal={Journal of educational technology systems},
  volume={49},
  number={1},
  pages={5--22},
  year={2020},
  publisher={Sage Publications Sage CA: Los Angeles, CA}
}

@inproceedings{ribeiro2016should,
  title={" Why should i trust you?" Explaining the predictions of any classifier},
  author={Ribeiro, Marco Tulio and Singh, Sameer and Guestrin, Carlos},
  booktitle={Proceedings of the 22nd ACM SIGKDD international conference on knowledge discovery and data mining},
  pages={1135--1144},
  year={2016}
}

@article{leo2021offline,
  title={From offline to online learning: A qualitative study of challenges and opportunities as a response to the COVID-19 pandemic in the UAE higher education context},
  author={Leo, Shirley and Alsharari, Nizar Mohammad and Abbas, Jainambu and Alshurideh, Muhammad Turki},
  journal={The effect of coronavirus disease (COVID-19) on business intelligence},
  pages={203--217},
  year={2021},
  publisher={Springer}
}

@article{shen2024survey,
  title={A Survey of Knowledge Tracing: Models, Variants, and Applications},
  author={Shen, Shuanghong and Liu, Qi and Huang, Zhenya and Zheng, Yonghe and Yin, Minghao and Wang, Minjuan and Chen, Enhong},
  journal={IEEE Transactions on Learning Technologies},
  year={2024},
  publisher={IEEE}
}


@article{song2022survey,
  title={A survey on deep learning based knowledge tracing},
  author={Song, Xiangyu and Li, Jianxin and Cai, Taotao and Yang, Shuiqiao and Yang, Tingting and Liu, Chengfei},
  journal={Knowledge-Based Systems},
  volume={258},
  pages={110036},
  year={2022},
  publisher={Elsevier}
}

@article{ogange2018student,
  title={Student perceptions of the effectiveness of formative assessment in an online learning environment},
  author={Ogange, Betty Obura and Agak, John O and Okelo, Kevin Odhiambo and Kiprotich, Peter},
  journal={Open Praxis},
  volume={10},
  number={1},
  pages={29--39},
  year={2018}
}

@article{gikandi2011online,
  title={Online formative assessment in higher education: A review of the literature},
  author={Gikandi, Joyce Wangui and Morrow, Donna and Davis, Niki E},
  journal={Computers \& education},
  volume={57},
  number={4},
  pages={2333--2351},
  year={2011},
  publisher={Elsevier}
}

@article{ayu2020online,
  title={Online learning: Leading e-learning at higher education},
  author={Ayu, Mutiara},
  journal={The Journal of English Literacy Education: The Teaching and Learning of English as a Foreign Language},
  volume={7},
  number={1},
  pages={47--54},
  year={2020}
}

@inproceedings{jung2023language,
  title={Language Proficiency Enhanced Knowledge Tracing},
  author={Jung, Heeseok and Yoo, Jaesang and Yoon, Yohaan and Jang, Yeonju},
  booktitle={International Conference on Intelligent Tutoring Systems},
  pages={3--15},
  year={2023},
  organization={Springer}
}

@article{lundberg2017unified,
  title={A unified approach to interpreting model predictions},
  author={Lundberg, Scott M and Lee, Su-In},
  journal={Advances in neural information processing systems},
  volume={30},
  year={2017}
}

@inproceedings{wolf-etal-2020-transformers,
    title = "Transformers: State-of-the-Art Natural Language Processing",
    author = "Thomas Wolf and Lysandre Debut and Victor Sanh and Julien Chaumond and Clement Delangue and Anthony Moi and Pierric Cistac and Tim Rault and Rémi Louf and Morgan Funtowicz and Joe Davison and Sam Shleifer and Patrick von Platen and Clara Ma and Yacine Jernite and Julien Plu and Canwen Xu and Teven Le Scao and Sylvain Gugger and Mariama Drame and Quentin Lhoest and Alexander M. Rush",
    booktitle = "Proceedings of the 2020 Conference on Empirical Methods in Natural Language Processing: System Demonstrations",
    month = oct,
    year = "2020",
    address = "Online",
    publisher = "Association for Computational Linguistics",
    url = "https://www.aclweb.org/anthology/2020.emnlp-demos.6",
    pages = "38--45"
}

@article{wang2024pre,
  title={Pre-training Question Embeddings for Improving Knowledge Tracing with Self-supervised Bi-graph Co-contrastive Learning},
  author={Wang, Wentao and Ma, Huifang and Zhao, Yan and Li, Zhixin},
  journal={ACM Transactions on Knowledge Discovery from Data},
  volume={18},
  number={4},
  pages={1--20},
  year={2024},
  publisher={ACM New York, NY}
}

@article{li2024explainable,
  title={Explainable Few-shot Knowledge Tracing},
  author={Li, Haoxuan and Yu, Jifan and Ouyang, Yuanxin and Liu, Zhuang and Rong, Wenge and Li, Juanzi and Xiong, Zhang},
  journal={arXiv preprint arXiv:2405.14391},
  year={2024}
}

@article{van2008visualizing,
  title={Visualizing data using t-SNE.},
  author={Van der Maaten, Laurens and Hinton, Geoffrey},
  journal={Journal of machine learning research},
  volume={9},
  number={11},
  year={2008}
}

@inproceedings{zhang2021knowledge,
  title={Knowledge tracing models’ predictive performance when a student starts a skill},
  author={Zhang, Jiayi and Das, Rohini and Baker, Ryan and Scruggs, Richard},
  booktitle={Proceedings of the 14th International Conference on Educational Data Mining. EDM, Paris, France},
  pages={625--629},
  year={2021}
}

@article{slater2018degree,
  title={Degree of error in Bayesian knowledge tracing estimates from differences in sample sizes},
  author={Slater, Stefan and Baker, Ryan S},
  journal={Behaviormetrika},
  volume={45},
  number={2},
  pages={475--493},
  year={2018},
  publisher={Springer}
}

@article{khurana2023natural,
  title={Natural language processing: State of the art, current trends and challenges},
  author={Khurana, Diksha and Koli, Aditya and Khatter, Kiran and Singh, Sukhdev},
  journal={Multimedia tools and applications},
  volume={82},
  number={3},
  pages={3713--3744},
  year={2023},
  publisher={Springer}
}

@article{piech2015deep,
  title={Deep knowledge tracing},
  author={Piech, Chris and Bassen, Jonathan and Huang, Jonathan and Ganguli, Surya and Sahami, Mehran and Guibas, Leonidas J and Sohl-Dickstein, Jascha},
  journal={Advances in neural information processing systems},
  volume={28},
  year={2015}
}

@article{shi2022code,
  title={Code-DKT: A code-based knowledge tracing model for programming tasks},
  author={Shi, Yang and Chi, Min and Barnes, Tiffany and Price, Thomas},
  journal={arXiv preprint arXiv:2206.03545},
  year={2022}
}

@inproceedings{yu2024eckt,
  title={ECKT: Enhancing Code Knowledge Tracing via Large Language Models},
  author={Yu, Yang and Zhou, Yingbo and Zhu, Yaokang and Ye, Yutong and Chen, Liangyu and Chen, Mingsong},
  booktitle={Proceedings of the Annual Meeting of the Cognitive Science Society},
  volume={46},
  year={2024}
}

@article{kasurinen2009estimating,
  title={Estimating programming knowledge with Bayesian knowledge tracing},
  author={Kasurinen, Jussi and Nikula, Uolevi},
  journal={ACM SIGCSE Bulletin},
  volume={41},
  number={3},
  pages={313--317},
  year={2009},
  publisher={ACM New York, NY, USA}
}

@inproceedings{meliana2018adopting,
  title={Adopting Good-Learners' Paths in an Intelligent Tutoring System},
  author={Meliana, Selly and Nurjanah, Dade},
  booktitle={2018 IEEE International Conference on Teaching, Assessment, and Learning for Engineering (TALE)},
  pages={877--882},
  year={2018},
  organization={IEEE}
}

@inproceedings{rivers2016learning,
  title={Learning curve analysis for programming: Which concepts do students struggle with?},
  author={Rivers, Kelly and Harpstead, Erik and Koedinger, Kenneth R},
  booktitle={ICER},
  volume={16},
  pages={143--151},
  year={2016},
  organization={ACM}
}

@inproceedings{hosseini2017stereotype,
  title={Stereotype modeling for Problem-Solving performance predictions in MOOCs and traditional courses},
  author={Hosseini, Roya and Brusilovsky, Peter and Yudelson, Michael and Hellas, Arto},
  booktitle={Proceedings of the 25th Conference on User Modeling, Adaptation and Personalization},
  pages={76--84},
  year={2017}
}

@inproceedings{wang2017deep,
  title={Deep knowledge tracing on programming exercises},
  author={Wang, Lisa and Sy, Angela and Liu, Larry and Piech, Chris},
  booktitle={Proceedings of the fourth (2017) ACM conference on learning@ scale},
  pages={201--204},
  year={2017}
}

@article{lee2024language,
  title={Language Model Can Do Knowledge Tracing: Simple but Effective Method to Integrate Language Model and Knowledge Tracing Task},
  author={Lee, Unggi and Bae, Jiyeong and Kim, Dohee and Lee, Sookbun and Park, Jaekwon and Ahn, Taekyung and Lee, Gunho and Stratton, Damji and Kim, Hyeoncheol},
  journal={arXiv preprint arXiv:2406.02893},
  year={2024}
}

@misc{CSEDM2019,
  title = {2nd Educational Data Mining in Computer Science Education (CSEDM) Workshop},
  author = {{CSEDM Workshop}},
  year = {2019},
  month = {March},
  day = {5},
  note = {In conjunction with LAK 2019 at Arizona State University, Tempe AZ, USA},
  url = {https://sites.google.com/asu.edu/csedm-ws-lak-2019},
  howpublished = {\url{https://sites.google.com/asu.edu/csedm-ws-lak-2019}}
}

@misc{cm_codexglue_code2text_java,
  title = {CodeXGLUE Code2Text Java Dataset},
  author = {CM},
  year = {2023},
  howpublished = {\url{https://huggingface.co/datasets/CM/codexglue_code2text_java}},
  note = {Accessed on July 27, 2024}
}

@misc{cm_codexglue_code2text_python,
  title = {CodeXGLUE Code2Text Python Dataset},
  author = {CM},
  year = {2023},
  howpublished = {\url{https://huggingface.co/datasets/CM/codexglue_code2text_python}},
  note = {Accessed on July 27, 2024}
}

@inproceedings{yumetamath,
  title={MetaMath: Bootstrap Your Own Mathematical Questions for Large Language Models},
  author={Yu, Longhui and Jiang, Weisen and Shi, Han and Jincheng, YU and Liu, Zhengying and Zhang, Yu and Kwok, James and Li, Zhenguo and Weller, Adrian and Liu, Weiyang},
  booktitle={The Twelfth International Conference on Learning Representations},
  year={2024}
}

@misc{openai2024hello,
  author       = {OpenAI},
  title        = {Hello GPT-4o},
  howpublished = {\url{https://openai.com/index/hello-gpt-4o/}},
  year         = {2024},
  note         = {Accessed: 2024-05-13}
}

@inproceedings{cheng2022adaptkt,
  title={Adaptkt: A domain adaptable method for knowledge tracing},
  author={Cheng, Song and Liu, Qi and Chen, Enhong and Zhang, Kai and Huang, Zhenya and Yin, Yu and Huang, Xiaoqing and Su, Yu},
  booktitle={Proceedings of the Fifteenth ACM International Conference on Web Search and Data Mining},
  pages={123--131},
  year={2022}
}

@article{xie2024domain,
  title={Domain Generalizable Knowledge Tracing via Concept Aggregation and Relation-Based Attention},
  author={Xie, Yuquan and Yang, Wanqi and Wei, Jinyu and Yang, Ming and Gao, Yang},
  journal={arXiv preprint arXiv:2407.02547},
  year={2024}
}

@article{tang2024domain,
  title={Domain adaptive knowledge tracing},
  author={Tang, Yumeng and Yang, Wanqi and Xie, Yuquan and Yang, Ming},
  journal={International Journal of Machine Learning and Cybernetics},
  pages={1--14},
  year={2024},
  publisher={Springer}
}

@article{jung2024clst,
  title={CLST: Cold-Start Mitigation in Knowledge Tracing by Aligning a Generative Language Model as a Students' Knowledge Tracer},
  author={Jung, Heeseok and Yoo, Jaesang and Yoon, Yohaan and Jang, Yeonju},
  journal={arXiv preprint arXiv:2406.10296},
  year={2024}
}

@article{gururangan2020dont,
  title={Don't stop pretraining: Adapt language models to domains and tasks},
  author={Gururangan, Suchin and Marasovi{\'c}, Ana and Swayamdipta, Swabha and Lo, Kyle and Beltagy, Iz and Downey, Doug and Smith, Noah A},
  journal={arXiv preprint arXiv:2004.10964},
  year={2020}
}

@article{singhal2023towards,
    author = {Singhal, Karan and Tu, Tao and Gottweis, Juraj and Sayres, Rory and Wulczyn, Ellery and Hou, Le and Clark, Kevin and Pfohl, Stephen and Cole-Lewis, Heather and Neal, Darlene and Schaekermann, Mike and Wang, Amy and Amin, Mohamed and Lachgar, Sami and Mansfield, Philip and Prakash, Sushant and Green, Bradley and Dominowska, Ewa and Aguera y Arcas, Blaise and Tomasev, Nenad and Liu, Yun and Wong, Renee and Semturs, Christopher and Mahdavi, S. Sara and Barral, Joelle and Webster, Dale and Corrado, Greg S. and Matias, Yossi and Azizi, Shekoofeh and Karthikesalingam, Alan and Natarajan, Vivek},
    title = {Towards Expert-Level Medical Question Answering with Large Language Models},
    journal = {arXiv preprint arXiv:2305.09617},
    year = {2023} 
}

@article{wu2023bloom,
    author = {Wu, Shijie and Irsoy, Ozan and Lu, Steven and Dabravolski, Vadim and Dredze, Mark and Gehrmann, Sebastian and Kambadur, Prabhanjan and Rosenberg, David and Mann, Gideon},
    title = {BloombergGPT: A Large Language Model for Finance},
    journal = {arXiv preprint arXiv:2303.17564},
    year = {2023}
}

@article{labrak2024bio,
    author = {Labrak, Yanis and Bazoge, Adrien and Morin, Emmanuel and Gourraud, Pierre-Antoine and Rouvier, Mickael and Dufour, Richard},
    title = {BioMistral: A Collection of Open-Source Pretrained Large Language
Models for Medical Domains},
    journal = {arXiv preprint arXiv:2402.10373},
    year = {2024}
}

@article{messer2024automated,
  title={Automated grading and feedback tools for programming education: A systematic review},
  author={Messer, Marcus and Brown, Neil CC and K{\"o}lling, Michael and Shi, Miaojing},
  journal={ACM Transactions on Computing Education},
  volume={24},
  number={1},
  pages={1--43},
  year={2024},
  publisher={ACM New York, NY}
}

@article{keuning2018systematic,
  title={A systematic literature review of automated feedback generation for programming exercises},
  author={Keuning, Hieke and Jeuring, Johan and Heeren, Bastiaan},
  journal={ACM Transactions on Computing Education (TOCE)},
  volume={19},
  number={1},
  pages={1--43},
  year={2018},
  publisher={ACM New York, NY, USA}
}

@article{cheng2023effects,
  title={Effects of an automated programming assessment system on the learning performances of experienced and novice learners},
  author={Cheng, Li-Chen and Li, Wei and Tseng, Judy CR},
  journal={Interactive Learning Environments},
  volume={31},
  number={8},
  pages={5347--5363},
  year={2023},
  publisher={Taylor \& Francis}
}

@article{liu2022knowledge,
  title={Knowledge tracing: A bibliometric analysis},
  author={Liu, Tongxi},
  journal={Computers and Education: Artificial Intelligence},
  volume={3},
  pages={100090},
  year={2022},
  publisher={Elsevier}
}

@article{Feng2020CodeBERT,
    author = {Feng, Zhangyin and Guo, Daya and Tang, Duyu and Duan, Nan and Feng, Xiaocheng and Gong, Ming and Shou, Linjun and Qin, Bing and Liu, Ting and Jiang, Daxin and Zhou, Ming},
    title ={CodeBERT: A Pre-Trained Model for Programming and Natural Languages},
    journal = {arXiv preprint arXiv:2002.08155},
    year = {2020} 
}

@Misc{accelerate,
  title = {Accelerate: Training and inference at scale made simple, efficient and adaptable.},
  author = {Sylvain Gugger and Lysandre Debut and Thomas Wolf and Philipp Schmid and Zachary Mueller and Sourab Mangrulkar and Marc Sun and Benjamin Bossan},
  howpublished = {\url{https://github.com/huggingface/accelerate}},
  year = {2022}
}

@misc{loshchilov2019decoupledweightdecayregularization,
      title={Decoupled Weight Decay Regularization}, 
      author={Ilya Loshchilov and Frank Hutter},
      year={2019},
      eprint={1711.05101},
      archivePrefix={arXiv},
      primaryClass={cs.LG},
      url={https://arxiv.org/abs/1711.05101}, 
}

\begin{thebibliography}{44}
\providecommand{\natexlab}[1]{#1}

\bibitem[{Abdelrahman et~al.(2022)Abdelrahman, Abdelfattah, Wang, and Lin}]{abdelrahman2022dbe}
Ghodai Abdelrahman, Sherif Abdelfattah, Qing Wang, and Yu~Lin. 2022.
\newblock Dbe-kt22: A knowledge tracing dataset based on online student evaluation.
\newblock \emph{arXiv preprint arXiv:2208.12651}.

\bibitem[{Cheng et~al.(2023)Cheng, Li, and Tseng}]{cheng2023effects}
Li-Chen Cheng, Wei Li, and Judy~CR Tseng. 2023.
\newblock Effects of an automated programming assessment system on the learning performances of experienced and novice learners.
\newblock \emph{Interactive Learning Environments}, 31(8):5347--5363.

\bibitem[{Cheng et~al.(2022)Cheng, Liu, Chen, Zhang, Huang, Yin, Huang, and Su}]{cheng2022adaptkt}
Song Cheng, Qi~Liu, Enhong Chen, Kai Zhang, Zhenya Huang, Yu~Yin, Xiaoqing Huang, and Yu~Su. 2022.
\newblock Adaptkt: A domain adaptable method for knowledge tracing.
\newblock In \emph{Proceedings of the Fifteenth ACM International Conference on Web Search and Data Mining}, pages 123--131.

\bibitem[{Clark et~al.(2020)Clark, Luong, Le, and Manning}]{clark2020electra}
Kevin Clark, Minh-Thang Luong, Quoc~V Le, and Christopher~D Manning. 2020.
\newblock Electra: Pre-training text encoders as discriminators rather than generators.
\newblock \emph{arXiv preprint arXiv:2003.10555}.

\bibitem[{CM(2023{\natexlab{a}})}]{cm_codexglue_code2text_java}
CM. 2023{\natexlab{a}}.
\newblock Codexglue code2text java dataset.
\newblock \url{https://huggingface.co/datasets/CM/codexglue_code2text_java}.
\newblock Accessed on July 27, 2024.

\bibitem[{CM(2023{\natexlab{b}})}]{cm_codexglue_code2text_python}
CM. 2023{\natexlab{b}}.
\newblock Codexglue code2text python dataset.
\newblock \url{https://huggingface.co/datasets/CM/codexglue_code2text_python}.
\newblock Accessed on July 27, 2024.

\bibitem[{{CSEDM Workshop}(2019)}]{CSEDM2019}
{CSEDM Workshop}. 2019.
\newblock \href {https://sites.google.com/asu.edu/csedm-ws-lak-2019} {2nd educational data mining in computer science education (csedm) workshop}.
\newblock \url{https://sites.google.com/asu.edu/csedm-ws-lak-2019}.
\newblock In conjunction with LAK 2019 at Arizona State University, Tempe AZ, USA.

\bibitem[{Devlin et~al.(2018)Devlin, Chang, Lee, and Toutanova}]{devlin2018bert}
Jacob Devlin, Ming-Wei Chang, Kenton Lee, and Kristina Toutanova. 2018.
\newblock Bert: Pre-training of deep bidirectional transformers for language understanding.
\newblock \emph{arXiv preprint arXiv:1810.04805}.

\bibitem[{Feng et~al.(2020)Feng, Guo, Tang, Duan, Feng, Gong, Shou, Qin, Liu, Jiang, and Zhou}]{Feng2020CodeBERT}
Zhangyin Feng, Daya Guo, Duyu Tang, Nan Duan, Xiaocheng Feng, Ming Gong, Linjun Shou, Bing Qin, Ting Liu, Daxin Jiang, and Ming Zhou. 2020.
\newblock Codebert: A pre-trained model for programming and natural languages.
\newblock \emph{arXiv preprint arXiv:2002.08155}.

\bibitem[{Ghosh et~al.(2020)Ghosh, Heffernan, and Lan}]{ghosh2020context}
Aritra Ghosh, Neil Heffernan, and Andrew~S Lan. 2020.
\newblock Context-aware attentive knowledge tracing.
\newblock In \emph{Proceedings of the 26th ACM SIGKDD international conference on knowledge discovery \& data mining}, pages 2330--2339.

\bibitem[{Gugger et~al.(2022)Gugger, Debut, Wolf, Schmid, Mueller, Mangrulkar, Sun, and Bossan}]{accelerate}
Sylvain Gugger, Lysandre Debut, Thomas Wolf, Philipp Schmid, Zachary Mueller, Sourab Mangrulkar, Marc Sun, and Benjamin Bossan. 2022.
\newblock Accelerate: Training and inference at scale made simple, efficient and adaptable.
\newblock \url{https://github.com/huggingface/accelerate}.

\bibitem[{Gururangan et~al.(2020)Gururangan, Marasovi{\'c}, Swayamdipta, Lo, Beltagy, Downey, and Smith}]{gururangan2020dont}
Suchin Gururangan, Ana Marasovi{\'c}, Swabha Swayamdipta, Kyle Lo, Iz~Beltagy, Doug Downey, and Noah~A Smith. 2020.
\newblock Don't stop pretraining: Adapt language models to domains and tasks.
\newblock \emph{arXiv preprint arXiv:2004.10964}.

\bibitem[{He et~al.(2021)He, Gao, and Chen}]{he2021debertav3}
Pengcheng He, Jianfeng Gao, and Weizhu Chen. 2021.
\newblock Debertav3: Improving deberta using electra-style pre-training with gradient-disentangled embedding sharing.
\newblock \emph{arXiv preprint arXiv:2111.09543}.

\bibitem[{Hosseini et~al.(2017)Hosseini, Brusilovsky, Yudelson, and Hellas}]{hosseini2017stereotype}
Roya Hosseini, Peter Brusilovsky, Michael Yudelson, and Arto Hellas. 2017.
\newblock Stereotype modeling for problem-solving performance predictions in moocs and traditional courses.
\newblock In \emph{Proceedings of the 25th Conference on User Modeling, Adaptation and Personalization}, pages 76--84.

\bibitem[{Jung et~al.(2024)Jung, Yoo, Yoon, and Jang}]{jung2024clst}
Heeseok Jung, Jaesang Yoo, Yohaan Yoon, and Yeonju Jang. 2024.
\newblock Clst: Cold-start mitigation in knowledge tracing by aligning a generative language model as a students' knowledge tracer.
\newblock \emph{arXiv preprint arXiv:2406.10296}.

\bibitem[{Kasurinen and Nikula(2009)}]{kasurinen2009estimating}
Jussi Kasurinen and Uolevi Nikula. 2009.
\newblock Estimating programming knowledge with bayesian knowledge tracing.
\newblock \emph{ACM SIGCSE Bulletin}, 41(3):313--317.

\bibitem[{Keuning et~al.(2018)Keuning, Jeuring, and Heeren}]{keuning2018systematic}
Hieke Keuning, Johan Jeuring, and Bastiaan Heeren. 2018.
\newblock A systematic literature review of automated feedback generation for programming exercises.
\newblock \emph{ACM Transactions on Computing Education (TOCE)}, 19(1):1--43.

\bibitem[{Labrak et~al.(2024)Labrak, Bazoge, Morin, Gourraud, Rouvier, and Dufour}]{labrak2024bio}
Yanis Labrak, Adrien Bazoge, Emmanuel Morin, Pierre-Antoine Gourraud, Mickael Rouvier, and Richard Dufour. 2024.
\newblock Biomistral: A collection of open-source pretrained large language models for medical domains.
\newblock \emph{arXiv preprint arXiv:2402.10373}.

\bibitem[{Lan et~al.(2019)Lan, Chen, Goodman, Gimpel, Sharma, and Soricut}]{lan2019albert}
Zhenzhong Lan, Mingda Chen, Sebastian Goodman, Kevin Gimpel, Piyush Sharma, and Radu Soricut. 2019.
\newblock Albert: A lite bert for self-supervised learning of language representations.
\newblock \emph{arXiv preprint arXiv:1909.11942}.

\bibitem[{Lee et~al.(2024)Lee, Bae, Kim, Lee, Park, Ahn, Lee, Stratton, and Kim}]{lee2024language}
Unggi Lee, Jiyeong Bae, Dohee Kim, Sookbun Lee, Jaekwon Park, Taekyung Ahn, Gunho Lee, Damji Stratton, and Hyeoncheol Kim. 2024.
\newblock Language model can do knowledge tracing: Simple but effective method to integrate language model and knowledge tracing task.
\newblock \emph{arXiv preprint arXiv:2406.02893}.

\bibitem[{Liu et~al.(2019{\natexlab{a}})Liu, Huang, Yin, Chen, Xiong, Su, and Hu}]{liu2019ekt}
Qi~Liu, Zhenya Huang, Yu~Yin, Enhong Chen, Hui Xiong, Yu~Su, and Guoping Hu. 2019{\natexlab{a}}.
\newblock Ekt: Exercise-aware knowledge tracing for student performance prediction.
\newblock \emph{IEEE Transactions on Knowledge and Data Engineering}, 33(1):100--115.

\bibitem[{Liu(2022)}]{liu2022knowledge}
Tongxi Liu. 2022.
\newblock Knowledge tracing: A bibliometric analysis.
\newblock \emph{Computers and Education: Artificial Intelligence}, 3:100090.

\bibitem[{Liu et~al.(2019{\natexlab{b}})Liu, Ott, Goyal, Du, Joshi, Chen, Levy, Lewis, Zettlemoyer, and Stoyanov}]{liu2019roberta}
Yinhan Liu, Myle Ott, Naman Goyal, Jingfei Du, Mandar Joshi, Danqi Chen, Omer Levy, Mike Lewis, Luke Zettlemoyer, and Veselin Stoyanov. 2019{\natexlab{b}}.
\newblock Roberta: A robustly optimized bert pretraining approach.
\newblock \emph{arXiv preprint arXiv:1907.11692}.

\bibitem[{Liu et~al.(2024)Liu, Liu, Guo, Chen, Huang, Zhao, Tang, Luo, and Weng}]{liu2024xes3g5m}
Zitao Liu, Qiongqiong Liu, Teng Guo, Jiahao Chen, Shuyan Huang, Xiangyu Zhao, Jiliang Tang, Weiqi Luo, and Jian Weng. 2024.
\newblock Xes3g5m: A knowledge tracing benchmark dataset with auxiliary information.
\newblock \emph{Advances in Neural Information Processing Systems}, 36.

\bibitem[{Loshchilov and Hutter(2019)}]{loshchilov2019decoupledweightdecayregularization}
Ilya Loshchilov and Frank Hutter. 2019.
\newblock \href {https://arxiv.org/abs/1711.05101} {Decoupled weight decay regularization}.
\newblock \emph{Preprint}, arXiv:1711.05101.

\bibitem[{Meliana and Nurjanah(2018)}]{meliana2018adopting}
Selly Meliana and Dade Nurjanah. 2018.
\newblock Adopting good-learners' paths in an intelligent tutoring system.
\newblock In \emph{2018 IEEE International Conference on Teaching, Assessment, and Learning for Engineering (TALE)}, pages 877--882. IEEE.

\bibitem[{Messer et~al.(2024)Messer, Brown, K{\"o}lling, and Shi}]{messer2024automated}
Marcus Messer, Neil~CC Brown, Michael K{\"o}lling, and Miaojing Shi. 2024.
\newblock Automated grading and feedback tools for programming education: A systematic review.
\newblock \emph{ACM Transactions on Computing Education}, 24(1):1--43.

\bibitem[{Nakagawa et~al.(2019)Nakagawa, Iwasawa, and Matsuo}]{nakagawa2019graph}
Hiromi Nakagawa, Yusuke Iwasawa, and Yutaka Matsuo. 2019.
\newblock Graph-based knowledge tracing: modeling student proficiency using graph neural network.
\newblock In \emph{IEEE/WIC/ACM International Conference on Web Intelligence}, pages 156--163.

\bibitem[{OpenAI(2024)}]{openai2024hello}
OpenAI. 2024.
\newblock Hello gpt-4o.
\newblock \url{https://openai.com/index/hello-gpt-4o/}.
\newblock Accessed: 2024-05-13.

\bibitem[{Pandey and Karypis(2019)}]{pandey2019self}
Shalini Pandey and George Karypis. 2019.
\newblock A self-attentive model for knowledge tracing.
\newblock \emph{arXiv preprint arXiv:1907.06837}.

\bibitem[{Piech et~al.(2015)Piech, Bassen, Huang, Ganguli, Sahami, Guibas, and Sohl-Dickstein}]{piech2015deep}
Chris Piech, Jonathan Bassen, Jonathan Huang, Surya Ganguli, Mehran Sahami, Leonidas~J Guibas, and Jascha Sohl-Dickstein. 2015.
\newblock Deep knowledge tracing.
\newblock \emph{Advances in neural information processing systems}, 28.

\bibitem[{Rivers et~al.(2016)Rivers, Harpstead, and Koedinger}]{rivers2016learning}
Kelly Rivers, Erik Harpstead, and Kenneth~R Koedinger. 2016.
\newblock Learning curve analysis for programming: Which concepts do students struggle with?
\newblock In \emph{ICER}, volume~16, pages 143--151. ACM.

\bibitem[{Sanh et~al.(2019)Sanh, Debut, Chaumond, and Wolf}]{sanh2019distilbert}
Victor Sanh, Lysandre Debut, Julien Chaumond, and Thomas Wolf. 2019.
\newblock Distilbert, a distilled version of bert: smaller, faster, cheaper and lighter.
\newblock \emph{arXiv preprint arXiv:1910.01108}.

\bibitem[{Shen et~al.(2024)Shen, Liu, Huang, Zheng, Yin, Wang, and Chen}]{shen2024survey}
Shuanghong Shen, Qi~Liu, Zhenya Huang, Yonghe Zheng, Minghao Yin, Minjuan Wang, and Enhong Chen. 2024.
\newblock A survey of knowledge tracing: Models, variants, and applications.
\newblock \emph{IEEE Transactions on Learning Technologies}.

\bibitem[{Shi et~al.(2022)Shi, Chi, Barnes, and Price}]{shi2022code}
Yang Shi, Min Chi, Tiffany Barnes, and Thomas Price. 2022.
\newblock Code-dkt: A code-based knowledge tracing model for programming tasks.
\newblock \emph{arXiv preprint arXiv:2206.03545}.

\bibitem[{Singhal et~al.(2023)Singhal, Tu, Gottweis, Sayres, Wulczyn, Hou, Clark, Pfohl, Cole-Lewis, Neal, Schaekermann, Wang, Amin, Lachgar, Mansfield, Prakash, Green, Dominowska, Aguera~y Arcas, Tomasev, Liu, Wong, Semturs, Mahdavi, Barral, Webster, Corrado, Matias, Azizi, Karthikesalingam, and Natarajan}]{singhal2023towards}
Karan Singhal, Tao Tu, Juraj Gottweis, Rory Sayres, Ellery Wulczyn, Le~Hou, Kevin Clark, Stephen Pfohl, Heather Cole-Lewis, Darlene Neal, Mike Schaekermann, Amy Wang, Mohamed Amin, Sami Lachgar, Philip Mansfield, Sushant Prakash, Bradley Green, Ewa Dominowska, Blaise Aguera~y Arcas, Nenad Tomasev, Yun Liu, Renee Wong, Christopher Semturs, S.~Sara Mahdavi, Joelle Barral, Dale Webster, Greg~S. Corrado, Yossi Matias, Shekoofeh Azizi, Alan Karthikesalingam, and Vivek Natarajan. 2023.
\newblock Towards expert-level medical question answering with large language models.
\newblock \emph{arXiv preprint arXiv:2305.09617}.

\bibitem[{Sun et~al.(2020)Sun, Wang, Li, Feng, Tian, Wu, and Wang}]{sun2020ernie}
Yu~Sun, Shuohuan Wang, Yukun Li, Shikun Feng, Hao Tian, Hua Wu, and Haifeng Wang. 2020.
\newblock Ernie 2.0: A continual pre-training framework for language understanding.
\newblock In \emph{Proceedings of the AAAI conference on artificial intelligence}, volume~34, pages 8968--8975.

\bibitem[{Tang et~al.(2024)Tang, Yang, Xie, and Yang}]{tang2024domain}
Yumeng Tang, Wanqi Yang, Yuquan Xie, and Ming Yang. 2024.
\newblock Domain adaptive knowledge tracing.
\newblock \emph{International Journal of Machine Learning and Cybernetics}, pages 1--14.

\bibitem[{Wang et~al.(2017)Wang, Sy, Liu, and Piech}]{wang2017deep}
Lisa Wang, Angela Sy, Larry Liu, and Chris Piech. 2017.
\newblock Deep knowledge tracing on programming exercises.
\newblock In \emph{Proceedings of the fourth (2017) ACM conference on learning@ scale}, pages 201--204.

\bibitem[{Wu et~al.(2023)Wu, Irsoy, Lu, Dabravolski, Dredze, Gehrmann, Kambadur, Rosenberg, and Mann}]{wu2023bloom}
Shijie Wu, Ozan Irsoy, Steven Lu, Vadim Dabravolski, Mark Dredze, Sebastian Gehrmann, Prabhanjan Kambadur, David Rosenberg, and Gideon Mann. 2023.
\newblock Bloomberggpt: A large language model for finance.
\newblock \emph{arXiv preprint arXiv:2303.17564}.

\bibitem[{Xie et~al.(2024)Xie, Yang, Wei, Yang, and Gao}]{xie2024domain}
Yuquan Xie, Wanqi Yang, Jinyu Wei, Ming Yang, and Yang Gao. 2024.
\newblock Domain generalizable knowledge tracing via concept aggregation and relation-based attention.
\newblock \emph{arXiv preprint arXiv:2407.02547}.

\bibitem[{Yu et~al.(2024{\natexlab{a}})Yu, Jiang, Shi, Jincheng, Liu, Zhang, Kwok, Li, Weller, and Liu}]{yumetamath}
Longhui Yu, Weisen Jiang, Han Shi, YU~Jincheng, Zhengying Liu, Yu~Zhang, James Kwok, Zhenguo Li, Adrian Weller, and Weiyang Liu. 2024{\natexlab{a}}.
\newblock Metamath: Bootstrap your own mathematical questions for large language models.
\newblock In \emph{The Twelfth International Conference on Learning Representations}.

\bibitem[{Yu et~al.(2024{\natexlab{b}})Yu, Zhou, Zhu, Ye, Chen, and Chen}]{yu2024eckt}
Yang Yu, Yingbo Zhou, Yaokang Zhu, Yutong Ye, Liangyu Chen, and Mingsong Chen. 2024{\natexlab{b}}.
\newblock Eckt: Enhancing code knowledge tracing via large language models.
\newblock In \emph{Proceedings of the Annual Meeting of the Cognitive Science Society}, volume~46.

\bibitem[{Zhang et~al.(2017)Zhang, Shi, King, and Yeung}]{zhang2017dynamic}
Jiani Zhang, Xingjian Shi, Irwin King, and Dit-Yan Yeung. 2017.
\newblock Dynamic key-value memory networks for knowledge tracing.
\newblock In \emph{Proceedings of the 26th international conference on World Wide Web}, pages 765--774.

\end{thebibliography}
